 \documentclass[pmlr,twocolumn]{jmlr} 

\usepackage{booktabs}

\usepackage[load-configurations=version-1]{siunitx} 

\theorembodyfont{\upshape}
\theoremheaderfont{\scshape}
\theorempostheader{:}
\theoremsep{\newline}

\jmlrproceedings{PMLR}{}%
\jmlrvolume{ML4H Extended Abstract Arxiv Index}
\jmlryear{2020}
\jmlrsubmitted{2020}
\jmlrpublished{}
\jmlrworkshop{Machine Learning for Health (ML4H) 2020}

\title[Accounting for Affect in PLR]{Accounting for Affect in Pain Level Recognition}

\author{%
\Name{Md Taufeeq Uddin} \Email{mdtaufeeq@usf.edu}\\
\addr {University of South Florida, Tampa, FL, USA}
\AND
\Name{Shaun Canavan} \Email{scanavan@usf.edu}\\
\addr {University of South Florida, Tampa, FL, USA}
\AND
\Name{Ghada Zamzmi} \Email{alzamzmiga@nih.gov}\\
\addr {University of South Florida, Tampa, FL, USA \\  Current affiliation National Institutes of Health}
}

\editor{}

\begin{document}

\maketitle

\begin{abstract}
In this work, we address the importance of affect in automated pain assessment and the implications in real-world settings. To achieve this, we curate a new physiological dataset by merging the publicly available bioVid pain and emotion datasets. We then investigate pain level recognition on this dataset simulating participants' naturalistic affective behaviors. Our findings demonstrate that acknowledging affect in pain assessment is essential. We observe degradation in recognition performance when simulating the existence of affect to validate pain assessment models that do not account for it. Conversely, we observe a performance boost in recognition when we account for affect. 
\end{abstract}

\section{Introduction}
Automated pain assessment is an essential component in clinical settings to ensure prompt intervention and appropriate treatment. Depending on the context, relying on verbal methods such as pain scales and questionnaires \cite{williamson2005pain} for pain assessment may not be objective, reliable, actionable, and scalable \cite{egede2017cumulative}.
Further, babies and unconscious patients are unable to provide verbal responses, while verbal responses from a mentally impaired adult may be unreliable. To tackle this challenge, studies have been performed to automate pain assessment \cite{walter2014automatic, lotsch2018machine}. 

There has been significant progress in automated pain assessment using physiological data \cite{gruss2015pain, werner2014automatic}, visual data \cite{werner2016automatic, zhao2016facial}, and fusion of them \cite{egede2017fusing, fabiano2019emotion} for either detecting pain presence \cite{lucey2011painful, mansoor2018pain} or estimating pain intensity \cite{ martinez2017personalized, uddin2020multimodal, salekin2020first}. These methods achieved good performance when applied to existing publicly available datasets such as UNBC-Mcmaster Pain dataset \cite{lucey2011painful}, bioVid pain dataset \cite{walter2013biovid}, and MIntPain dataset \cite{haque2018deep}. 

These previous works are limited because they only compare pain to a neutral state (baseline). This can lead to the failure of these approaches in real-world settings where patients are likely to be in multiple affect states aside from being in pain or neutral. For example, it has been reported that hospitalized patients undergo a wide range of affects including anger, depression, anxiety in addition to the pain affect \cite{mirani2019frequency}.
These affect states can occur as a result of disease, stress, financial concerns, and unsatisfactory treatment \cite{mirani2019frequency}. Patients can also experience different positive affects, such as comfort or joy, as a result of interactions with family/caregivers \cite{eggins2014hospital}. 
These studies provide evidence that affect is relevant in pain assessment \cite{gatchel2007biopsychosocial}, and motivate us to consider affect states while developing pain assessment models. Such models would lead to a reliable and accurate assessment when applied to patients in clinical settings. 

In this paper, we propose to account for affect when assessing pain to provide a robust assessment in real-world settings given that the model is aware of other affective states. We curate a new pain-affect dataset we refer to as \textbf{bioVid pain-affect} (Section \ref{sec:dataset}). This allows us to build a reliable and robust pain assessment model that considers multiple affect states commonly experienced by patients (e.g., sadness, fear) in clinical settings. Our experimental results on the pain-affect dataset indicate that pain assessment models that do not take affect into consideration are likely to drift and make poor inferences when applied to real-world settings. 
In addition to reliable real-world assessment, the proposed modeling of pain allows us to adjust pain score/intensity based on underlying affect of the patient; e.g., it has been shown that there is altered pain sensitivity in patients who experience pre-existing anxiety and depression \cite{sullivan2017association,hermesdorf2016pain}. Finally, it can lead to the development of personalized treatment plans as it accounts for underlying states of each patient.

\section{bioVid Pain-Affect Dataset}
\label{sec:dataset}

We curate a new dataset by merging the publicly available bioVid pain \cite{walter2013biovid} and bioVid emotion \cite{zhang2016biovid} datasets. The pain dataset contains two sessions of recording participants' pain and baseline states. The emotion dataset contains one session of recording participants' discrete affects including amusement, anger, disgust, fear, and sadness. The data were collected in the same lab settings using similar equipment. Participants' age ranged from $19$ to $65$ years, and in total there are 91 subjects (45 female and 46 male) across both datasets. Further description of the datasets and stimuli for spontaneous pain and affect states can be found in Appendix A, Section 2. 

To merge the datasets, we selected the common subjects that appear in both the pain and emotion datasets (82 subjects in total). We then removed $20$ participants from the merged dataset, who did not show pain responses during the pain elicitation process \cite{werner2017analysis}, resulting in $62$ participants ($33$ female and $29$ male). Merging of the datasets resulted in a new age range of $[20, 65]$ with a median age of $36$. The merged bioVid pain-affect dataset contains three biopotential signals: skin conductance level or electrodermal activity (\textbf{EDA}), electrocardiogram (\textbf{ECG}), electromyogram (\textbf{EMG}) of trapezius muscle.

\section{Pain Assessment Model}

\subsection{Problem Formulation}
Our work aims to recognize pain level while taking into account affect states that exist in real-world settings such as anger and anxiety. Recall that we combine discrete affect states (e.g., sadness, fear) to one affect state named \textbf{A}. We hypothesize that the consideration of different affect while assessing pain is necessary to obtain accurate results, and for transparency. To do so, we propose to incorporate affect in PL recognition model as a category (e.g., $A$ in our studied dataset) along with multiple pain levels (e.g., \textbf{LLP} - low-level pain, \textbf{HLP} - high-level pain in our studied dataset; see Appendix A Section 2 for details). Depending on the context, baseline (\textbf{BL}) category could be incorporated in affect category as BL is likely to be a neutral/relaxed affect state, and for some subjects, relaxed states could be other positive or negative affect states (e.g., amusement, fear) depending on the situation, environment, or time. 

\vspace{-5mm}
\subsection{Feature Representation}
To create a feature vector for a given sample, we downsampled the biopotential signals by computing the moving average using a sliding window with $80\%$ overlap. Before downsampling the signals, we used a Savitzky–Golay filter \cite{press1990savitzky} to remove noise from the raw signals. We also normalized each signal in the range $[0, 1]$ using min-max normalization. We then created a normalized feature vector as $f_m = [c_1, c_2, \dots, c_n]$, where $c$ and $n$ are the downsampled signal frames/components and length of downsampled signal, respectively.  

\subsection{Case studies}
    \textbf{Case -1.} We explore the current naive PL recognition (\textbf{PLR}) models, in which we use only BL and pain data during training, validation and testing. 

    \textbf{Case 0.} We simulate real-world settings on the naive PLR model. As we discussed before, a person is likely to experience pain, baseline, and other affects in real-world settings. Hence, we reproduce the PLR model developed in case -1, and tested the model assuming the affect exists in our testing dataset.
    
    \textbf{Case 5.} This case is based on our proposal (Section 3.1). In this case, we acknowledge during the development of PLR model and during testing in real-world settings, that a person would experience affect (e.g., sadness, fear), baseline, and multiple levels of pain. In our training, validation, and testing datasets, we used affect and pain data. In the affect data, we incorporate the BL state assuming it is part of affect state. 
    
    \textbf{Case 6.} This case is a variation of case $5$ in which we remove the baseline data from other affect data. This case allows us to investigate the performance of PLR model in absence of baseline, and in presence of other affect. Also, in this case, we also investigate the impact of a balanced class distribution among classes (e.g., LLP, HLP, $A$) in the training dataset. Contrary to cases $-1, 0, 5$, where we preserve the training dataset uniformity in terms of number of examples and class proportion to ensure fair evaluation.

\section{Experiments and Analysis}

\begin{figure*}
\centering
\includegraphics[width=0.24\textwidth]{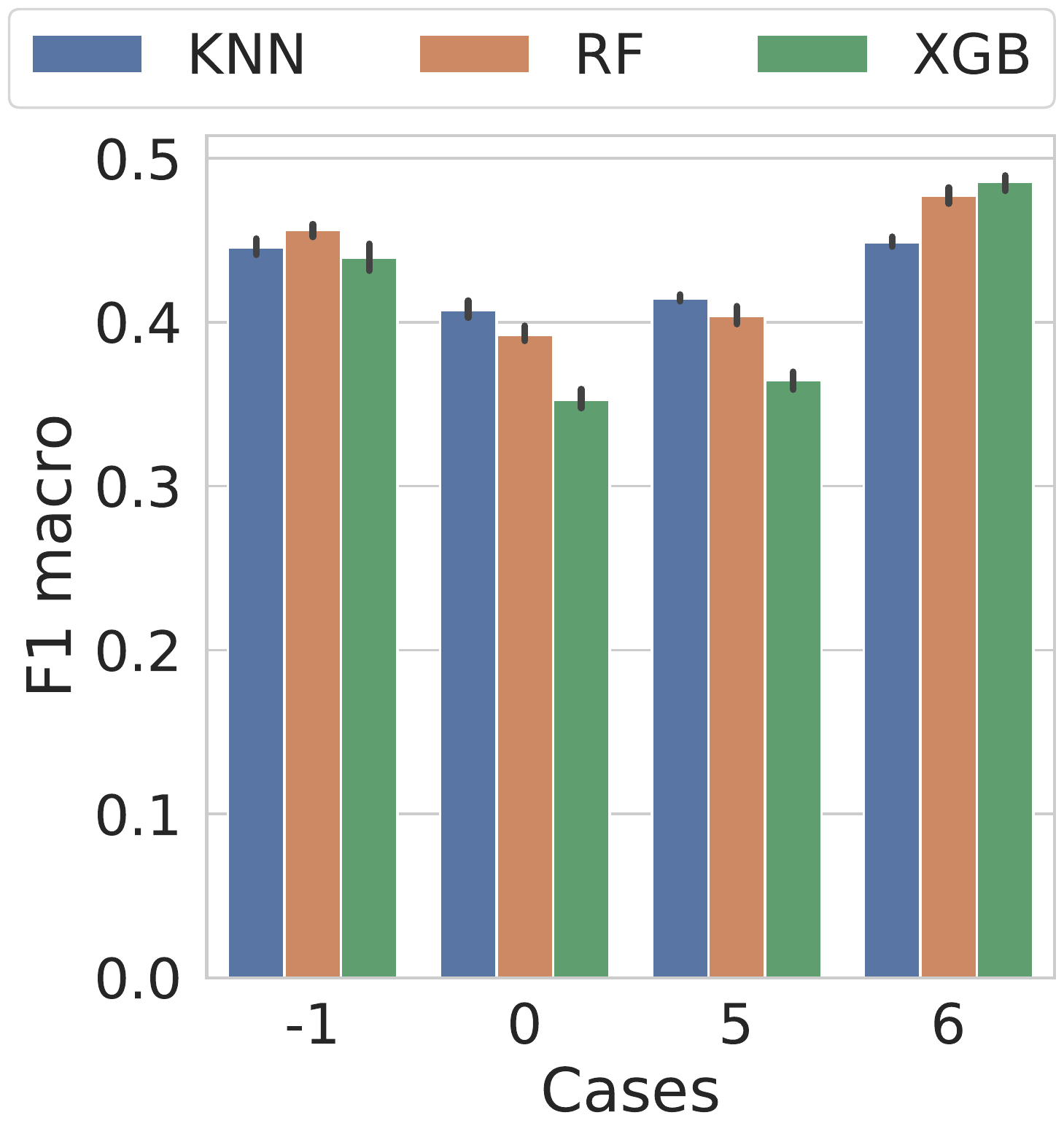}
\hfill
\includegraphics[width=0.24\textwidth]{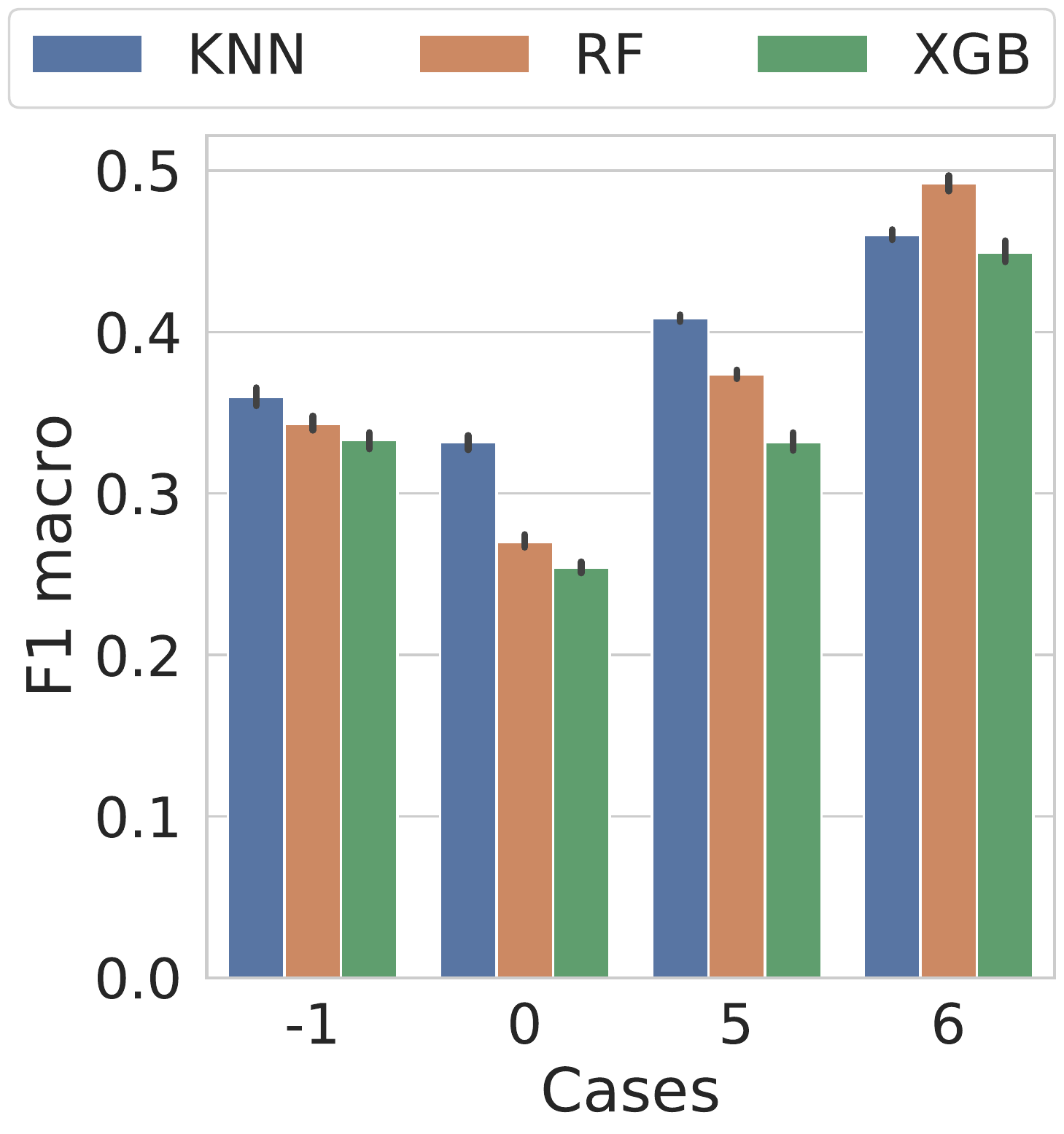}
\hfill
\includegraphics[width=0.24\textwidth]{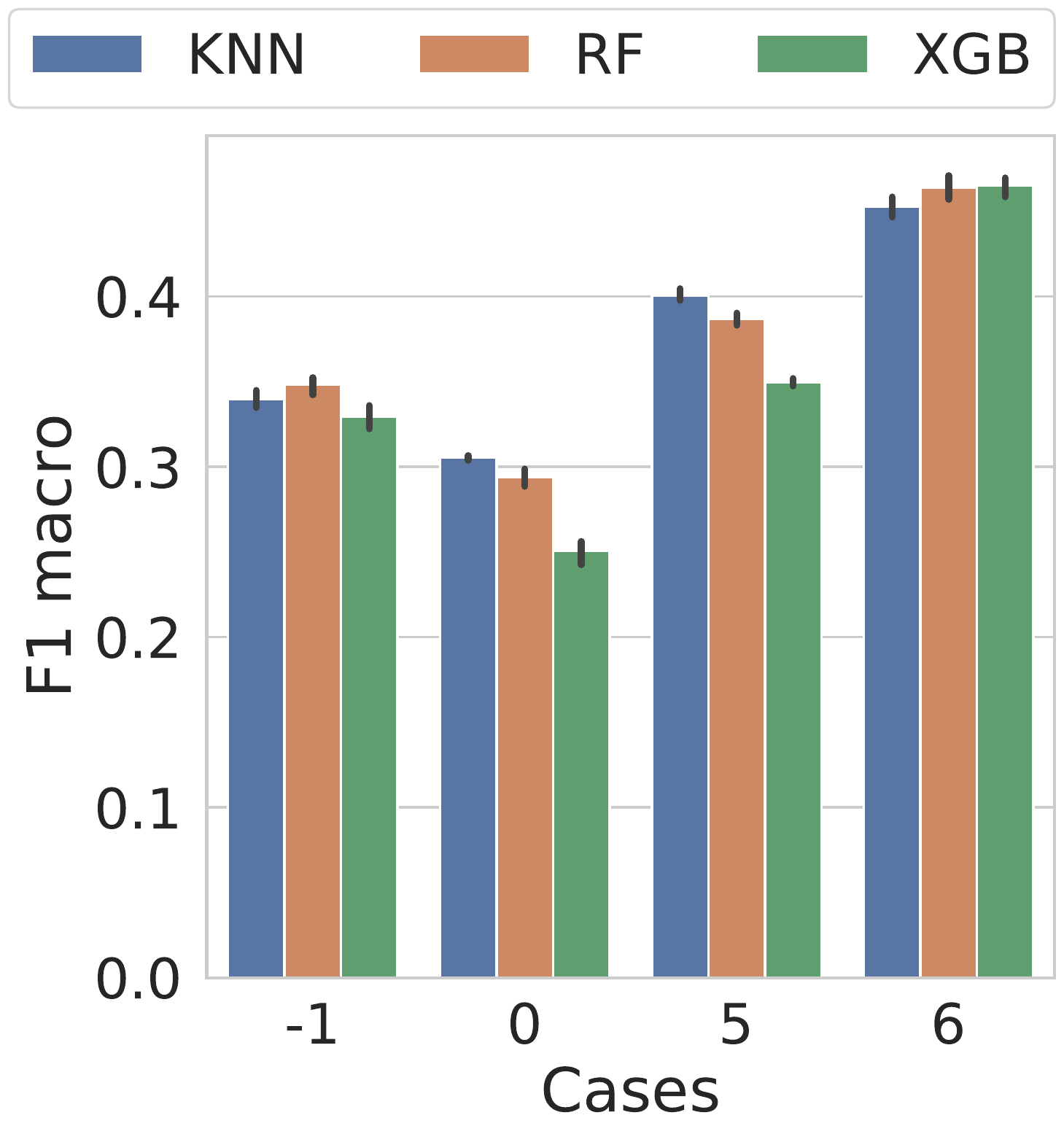}
\hfill
\includegraphics[width=0.24\textwidth]{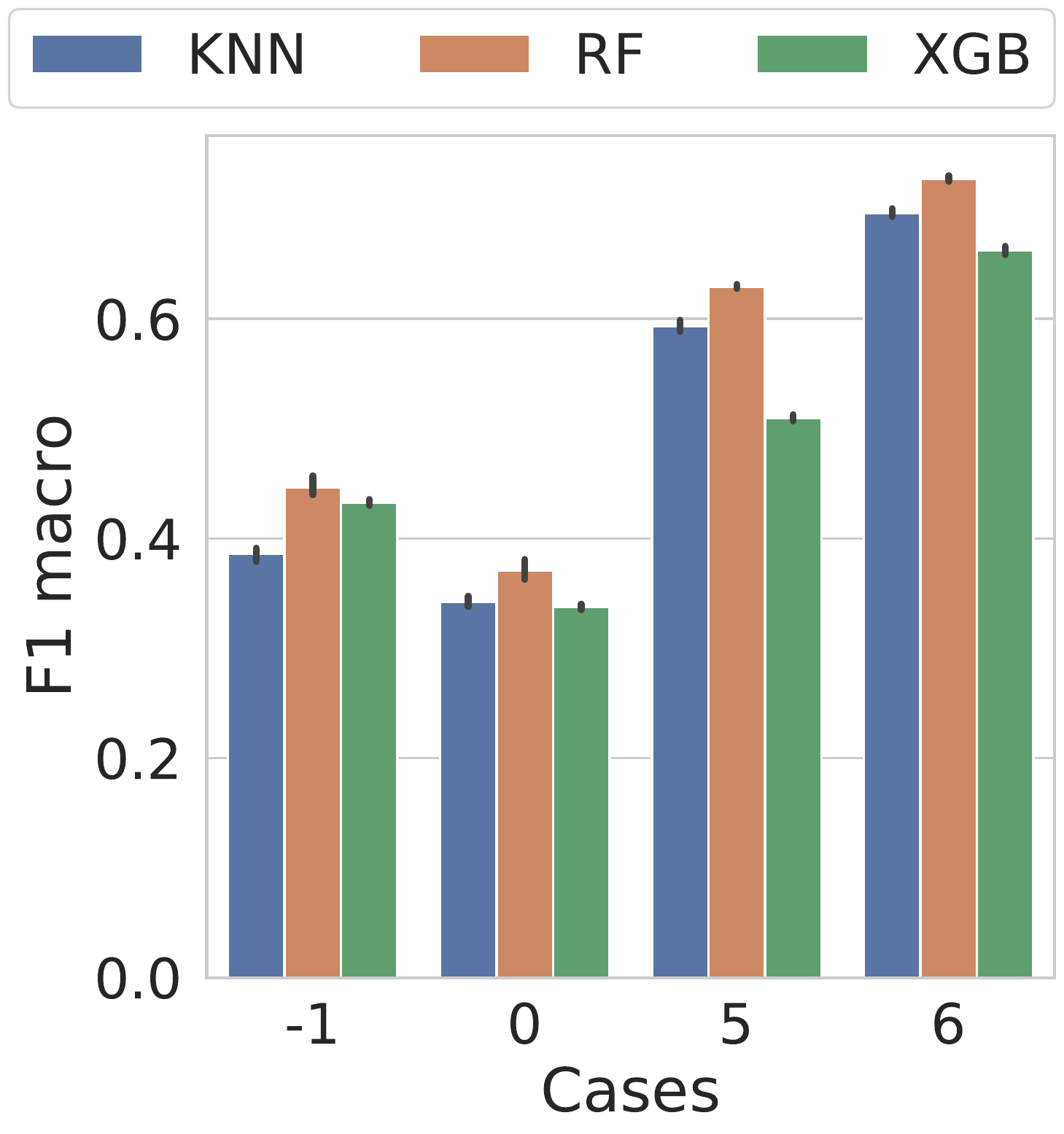}
\caption{PLR model performance on \textbf{known participants} (we assumed participants are likely to known) in investigated cases. From left to right, figures represent results on \textbf{EDA, ECG, EMG,} and \textbf{EDA + ECG + EMG} modalities, respectively.}
\label{fig:genRand}
\vspace{-5mm}
\end{figure*}

\textbf{Evaluation setup.} To validate the model, we assumed that participants were known during validation. Hence, we randomly split the whole dataset into $70\%$ training and $30\%$ test data, and we ran the experiments $5$ times with different seeds. F1 macro was used to report the model performance.

\textbf{Modalities and Classifiers}. To investigate the performance of PLR models in above mentioned cases, we built the models in unimodel and multimodel settings, i.e., we trained and tested PLR model on EDA, ECG, EMG separately, and their combination (EDA + ECG + EMG). We experimented with three well-known classification algorithms: K Nearest Neighbors (\textbf{KNN}, $k=5$) \cite{devroye2013probabilistic}, Random Forests (\textbf{RF}, tree estimator = $750$) \cite{breiman2001random}, and Extreme Gradient Boosting (\textbf{XGB}, tree estimator = $750$ and learning rate = $0.1$.) \cite{chen2016xgboost}. 

\textbf{Case Studies Evaluation}. Recall that in case $-1$ and case $0$, we trained the naive PLR model, i.e. model does not know about affect (e.g., anger, sadness). Also, training data size and class proportion were kept uniform in case $-1, 0,$ and $5$ for fair evaluation. When we tested the naive PLR model in case $0$ (simulated the existence of affects; see Figure \ref{fig:genRand}, the model showed drift in terms of recognition rate compared to the results obtained in case $-1$. This can be attributed to the PLR model not being aware of the affect data during training. Some evidence behind this issue could be found from Figures \ref{fig:sampDist} and \ref{fig:lSpace} (Appendix A). Since we are missing out on affect data in the naive modeling case, it is hard for the naive PLR model to generalize or adapt to the affect data. More precisely, during inference in simulated natural world, the trained naive PL recognition model seemed to treat the affect data as unknown data. 
 
In case $5$, we take the affect into account in our PLR model. Note that test dataset is exactly same in both case $5$ and case $0$. As shown in Figure \ref{fig:genRand}, the model showed significant improvement in terms of recognition performance compared to case $0$. This can be explained, in part, due to the affect data distribution being known to the proposed PLR model. Finally, recall that case $6$ was designed to investigate the influence of a balanced training dataset in terms of class proportion. Also, to evaluate the proposed model in the absence of BL data in both training and test dataset. In this case (Figure \ref{fig:genRand}), we are seeing an improvement in recognition performance, which can be explained by the even distribution of LLP, HLP, and A categories in training dataset.

Section 4 in Appendix A presents further experiments, evaluation, and analysis of PLR models using different experimental setup.

\section{Conclusion and Future Work}

This work investigated the necessity of including non-pain affect in pain level recognition, which is a major limitation in current PLR approaches. Using just a baseline during PLR modeling (case $-1$ in Section 3.3) produced reasonable results in a controlled environment. However, as evidenced from our experiments (Figures \ref{fig:genRand} above, and Figures \ref{fig:genSubOut}, \ref{fig:person}, \ref{fig:genRandWDemog}, \ref{fig:genSubOutWDemog} in Appendix A), these naive models fail in natural settings as people experience a diverse set of affects apart from neutral and pain. Hence, we conclude that incorporating affect in PLR model is crucial.

The main limitation of this study is the utilization of only physiological signals. In the future, we will explore behavioral data such as face images, facial expressiveness dynamics \cite{uddin2020quantified} and facial action units \cite{ekman1997face} along with the combination of physiological and behavioral data.

\clearpage
\bibstyle{unsrt}
\bibliography{IEEEexample}

\clearpage
\section*{Appendix A}\label{apd:first}

\begin{figure*}[b]
  \centering
  \includegraphics[width=.33\textwidth]{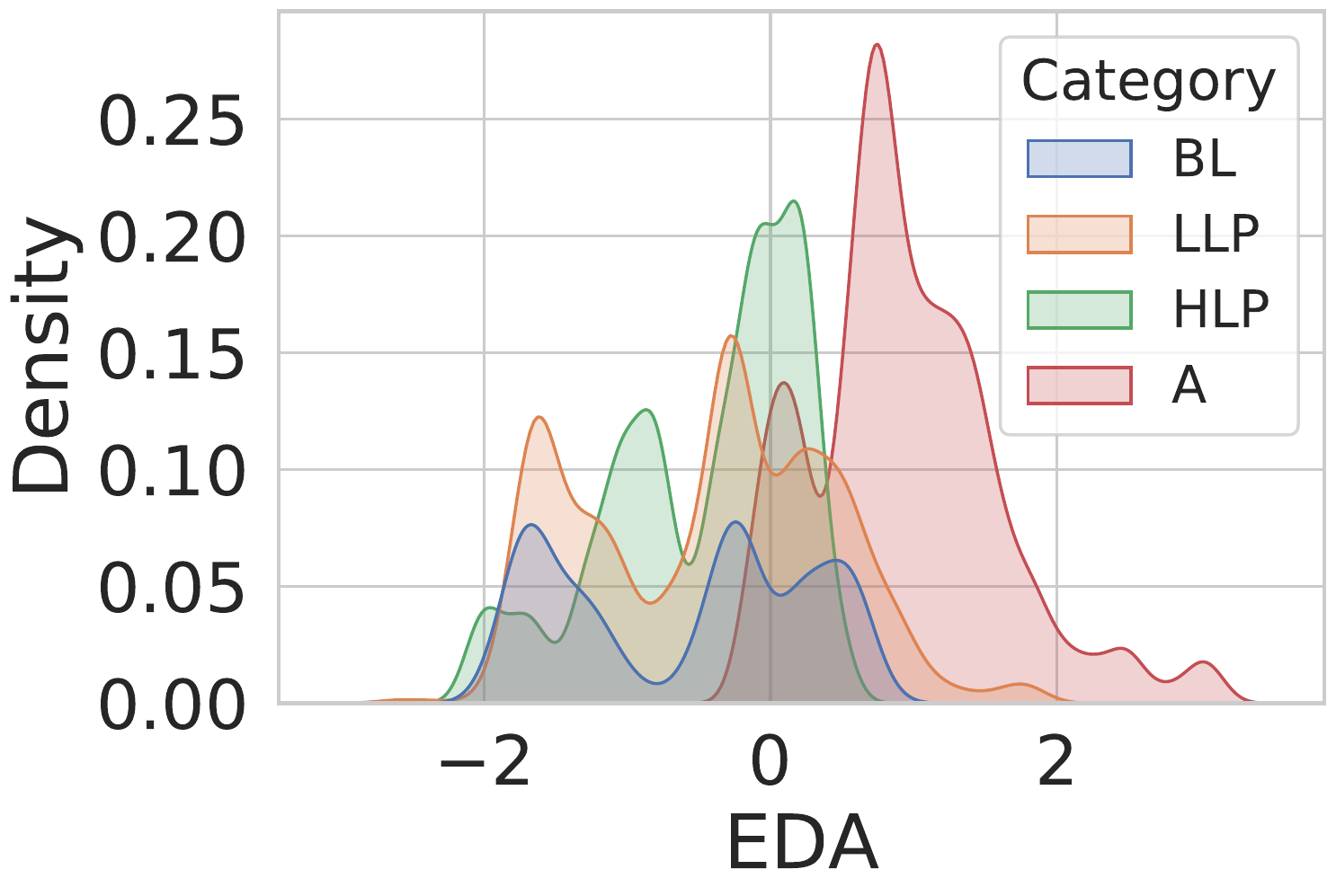}\hfill
  \includegraphics[width=.33\textwidth]{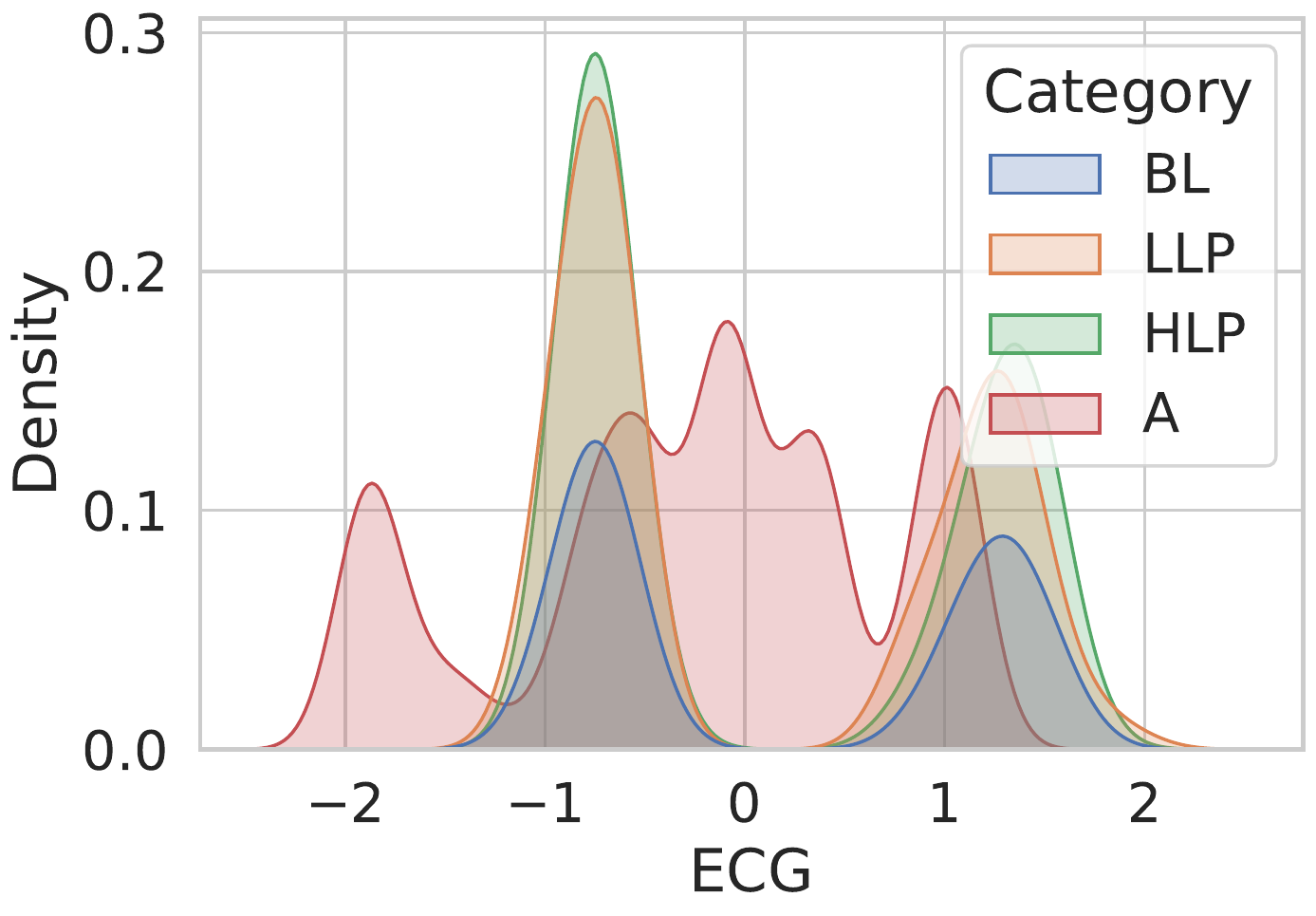}\hfill
  \includegraphics[width=.33\textwidth]{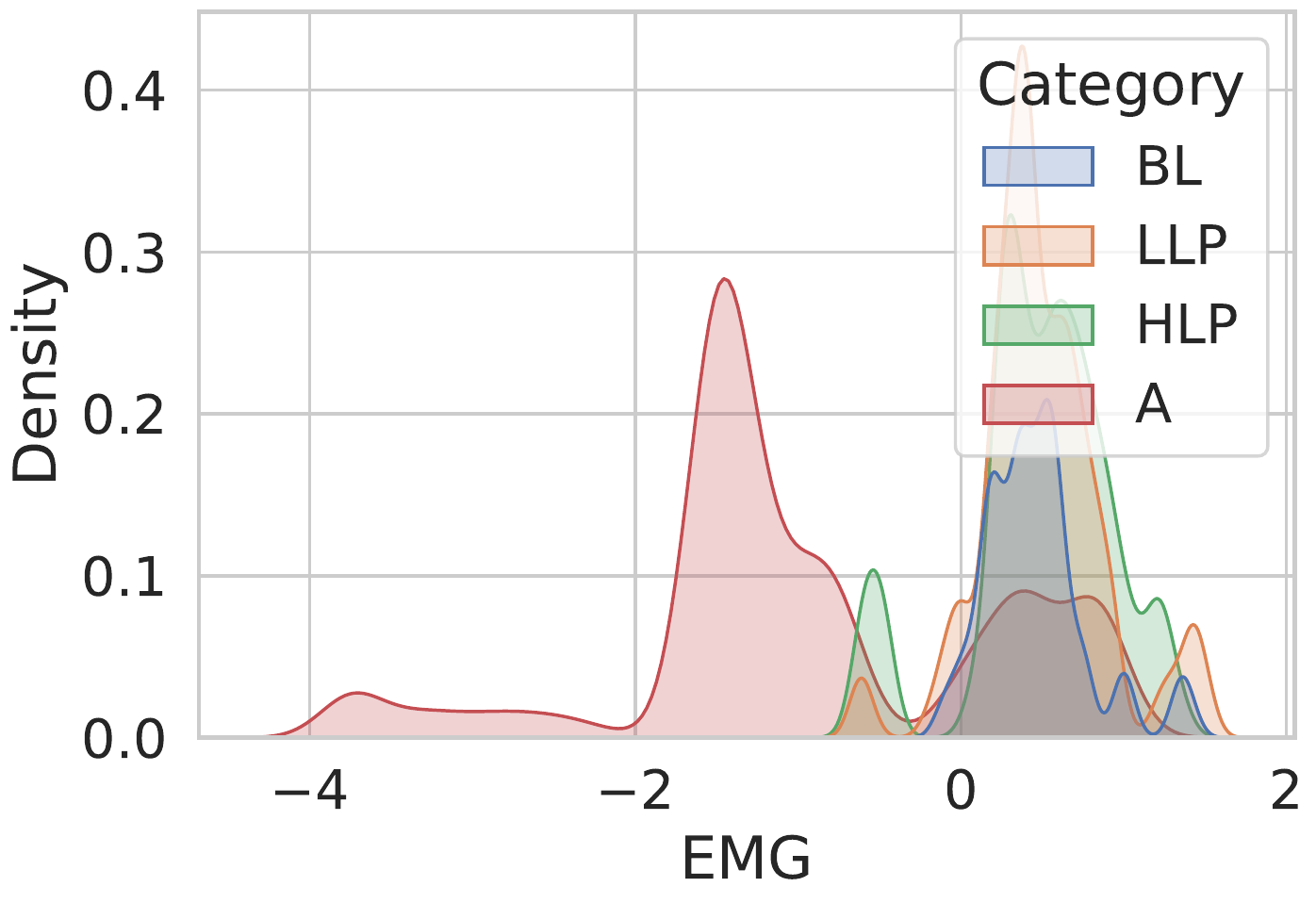}\hfill
  \caption{Distribution of physiological signals depicting affective states. Here, standardized (zero-mean and unit variance) data were sampled from one participant. We can see that a subject could be in BL, LLP, HLP, and A states in different moments of time in natural world.}
  \label{fig:sampDist}
  \vspace{-5mm}
\end{figure*}

\begin{figure*}
\centering
\includegraphics[width=0.32\textwidth]{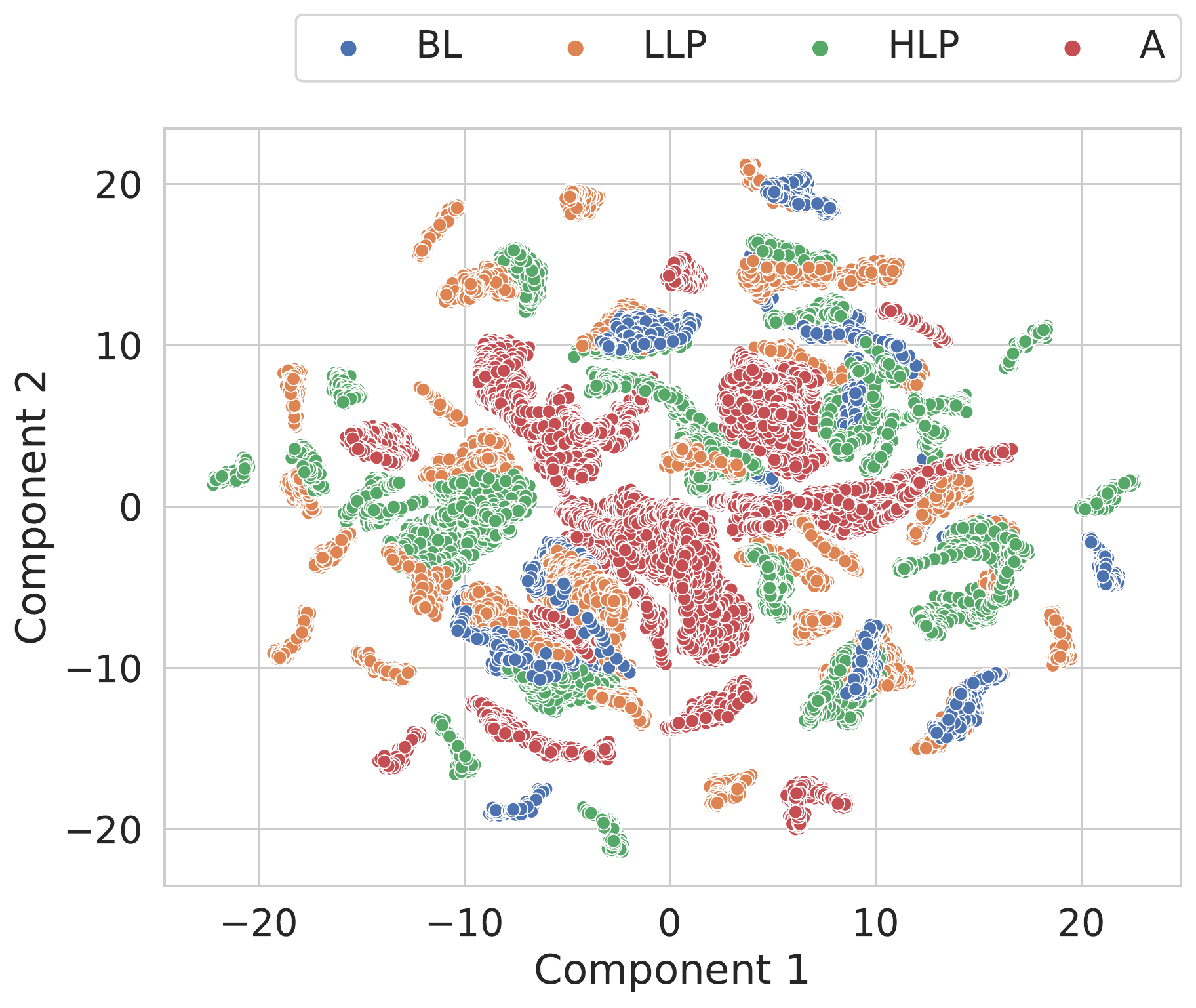}%
\label{ls1}
\hfill
\includegraphics[width=0.32\textwidth]{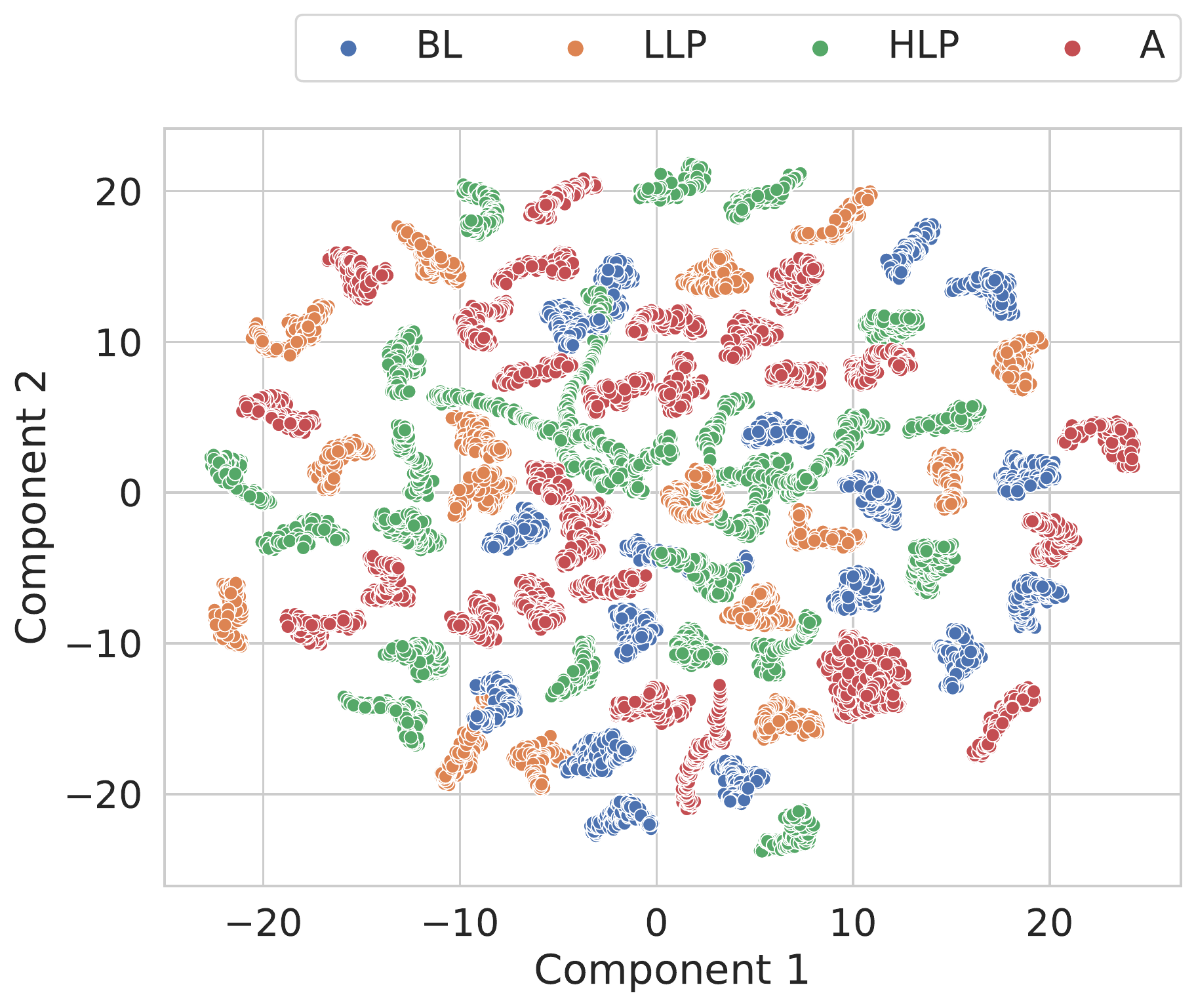}%
\label{ls2}
\hfill
\includegraphics[width=0.32\textwidth]{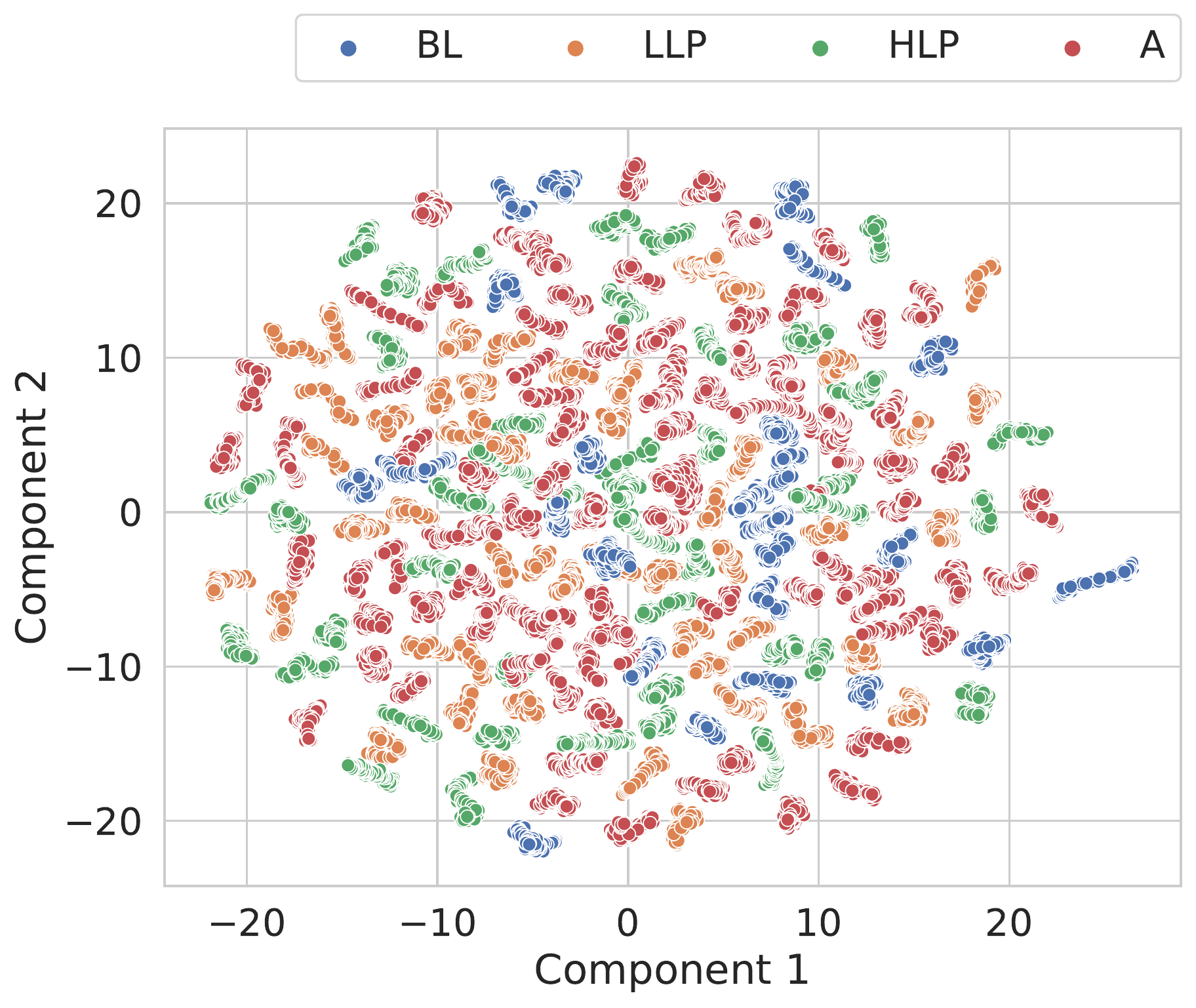}%
\label{ls3}
\caption{Latent space representation of physiological data depicting affective states, using UMAP (uniform manifold approximation and projection) algorithm \cite{mcinnes2018umap}. From left to right, figures depict \textbf{person-specific, demography-specific} and \textbf{general} scenarios. A grid of values were used to tune the major parameters (nearest neighbors = $[2, 5, 10, 20, 50, 100, 200]$, and minimum distance = $[0.0, 0.1, 0.25, 0.5, 0.8, 0.99]$) of UMAP. In person-specific scenario, data were sampled from one participant; demography-specific scenario, data were sampled from one specific demographic group (see footnote 1); general scenario, data were sampled from all $62$ participants. }
\label{fig:lSpace}
\vspace{-5mm}
\end{figure*}

\begin{figure*}
\centering
\includegraphics[width=0.24\textwidth]{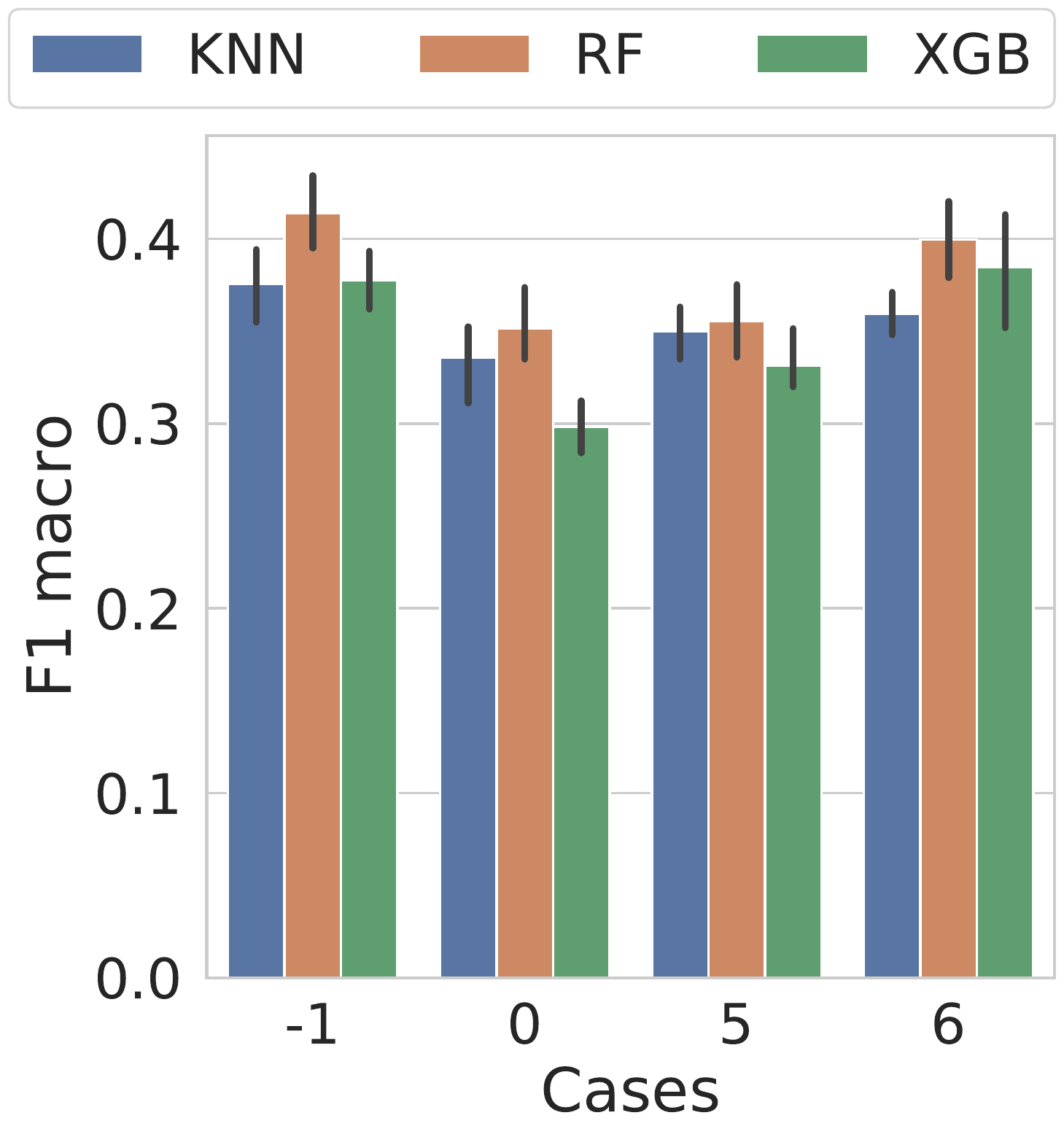}%
\hfill
\includegraphics[width=0.24\textwidth]{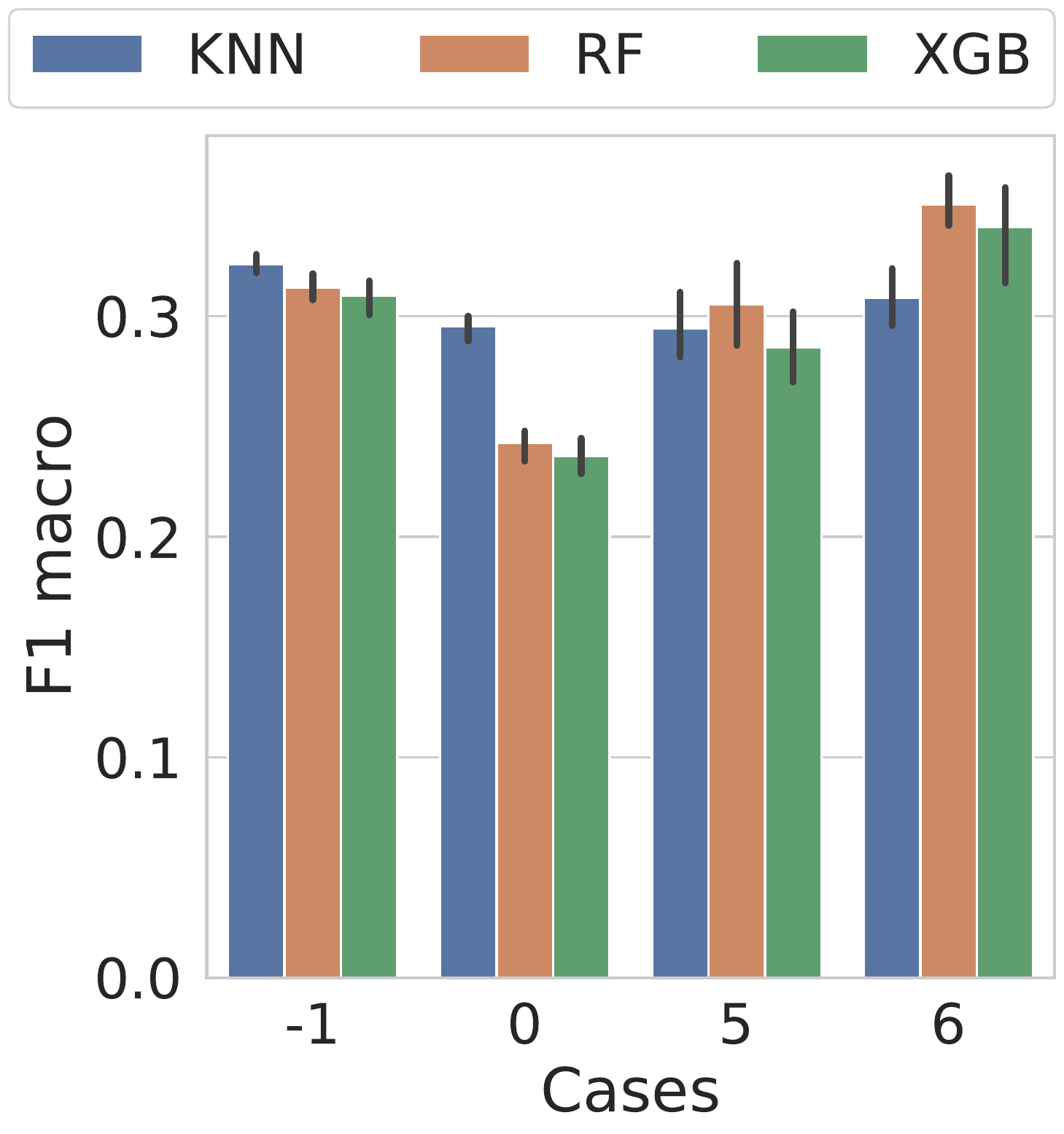}%
\hfill
\includegraphics[width=0.24\textwidth]{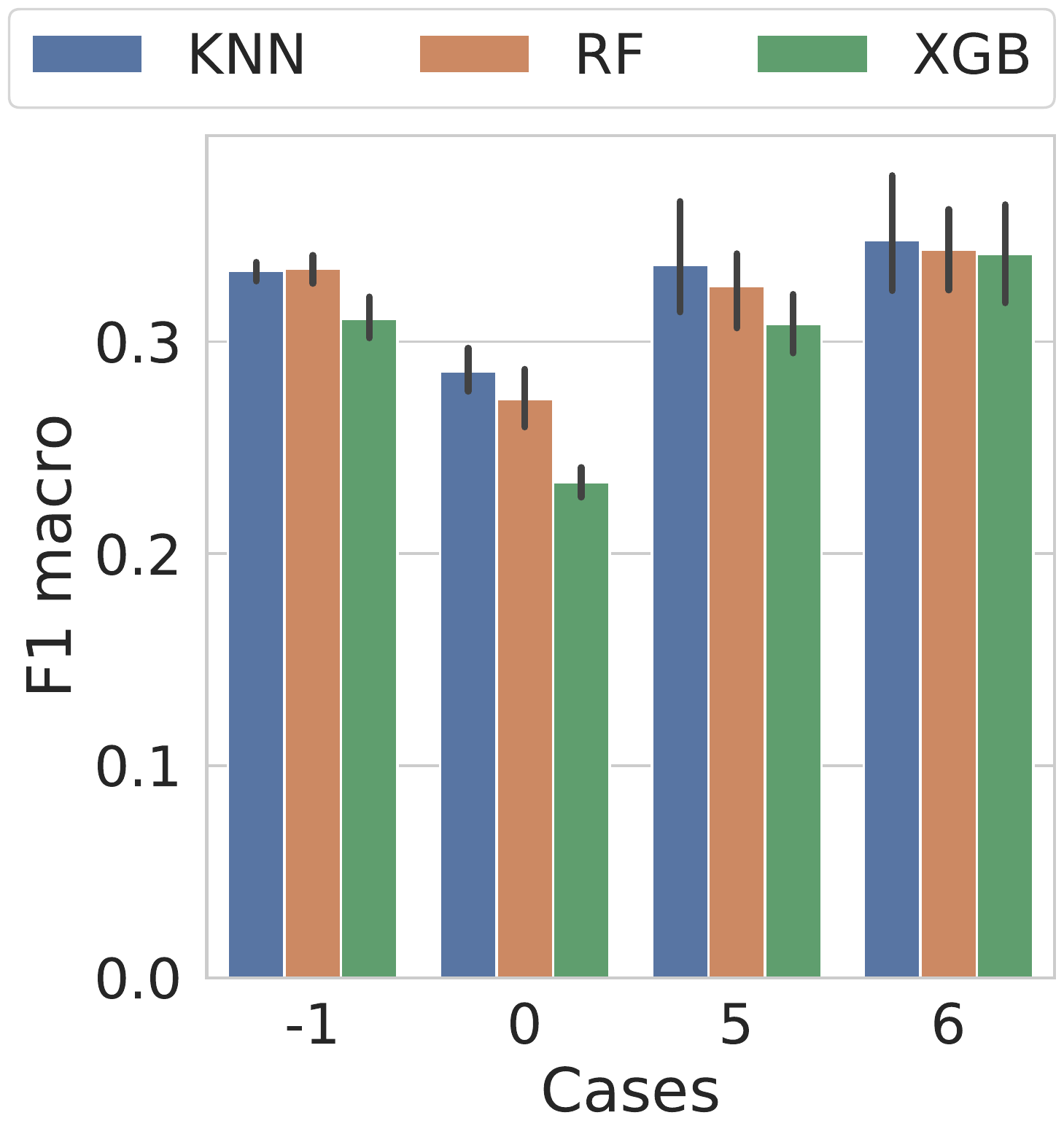}%
\hfill
\includegraphics[width=0.24\textwidth]{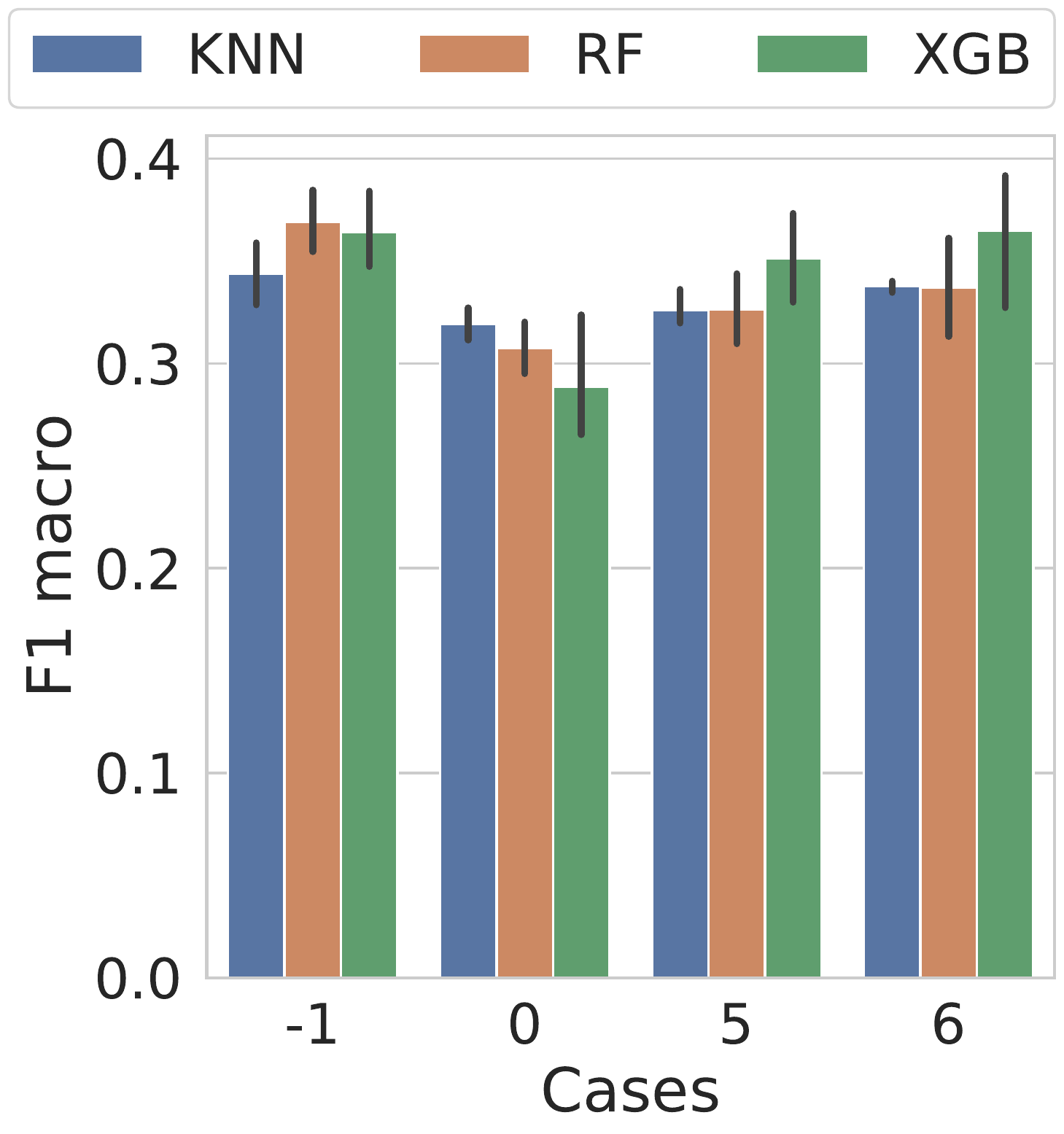}%
\caption{PLR model performance on \textbf{unknown participants} in investigated cases. From left to right, figures represent results on \textbf{EDA, ECG, EMG,} and \textbf{EDA + ECG + EMG} modalities, respectively.}
\label{fig:genSubOut}
\vspace{-5mm}
\end{figure*}

\begin{figure*}
\centering
\includegraphics[width=0.24\textwidth]{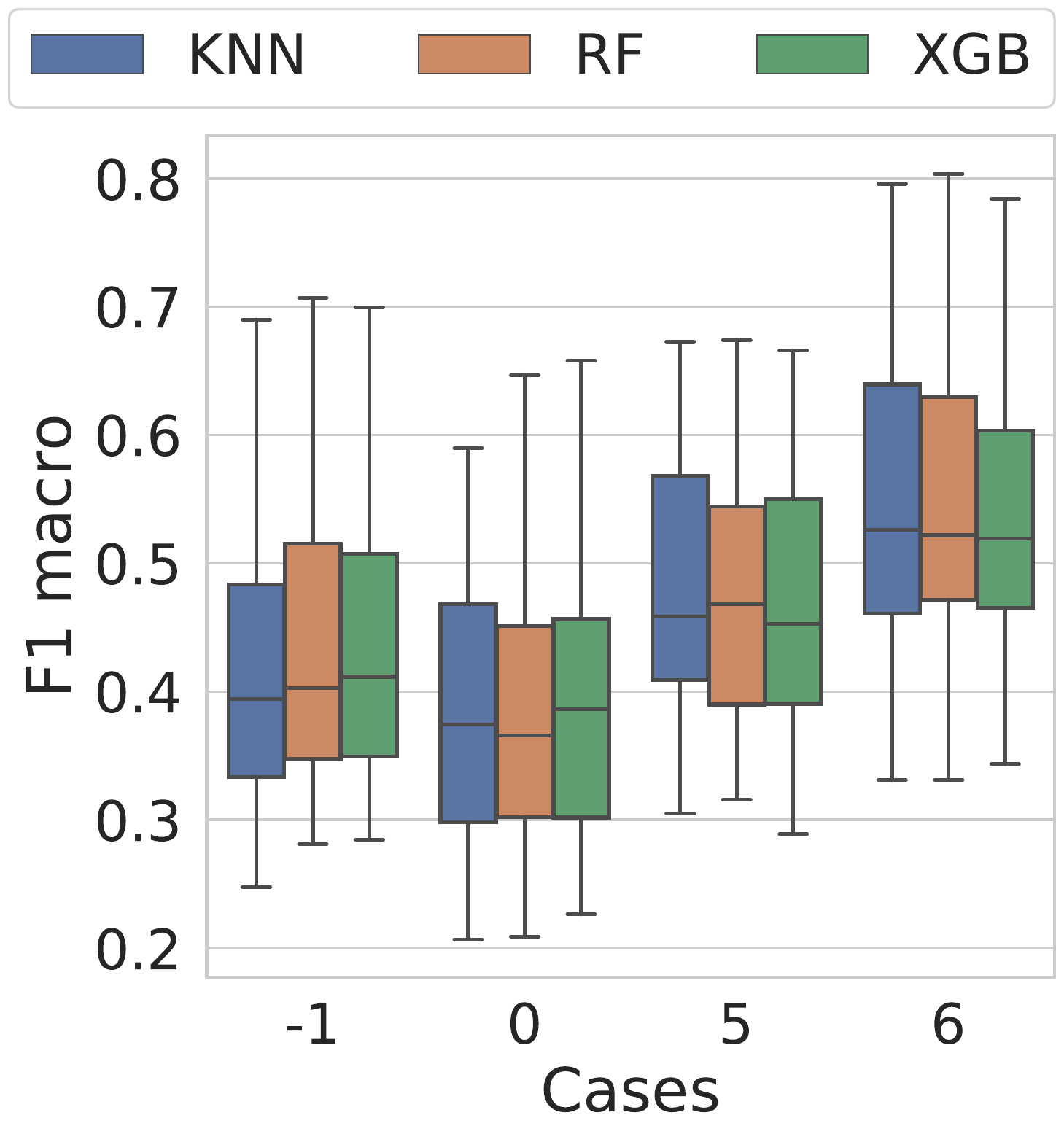}%
\hfill
\includegraphics[width=0.24\textwidth]{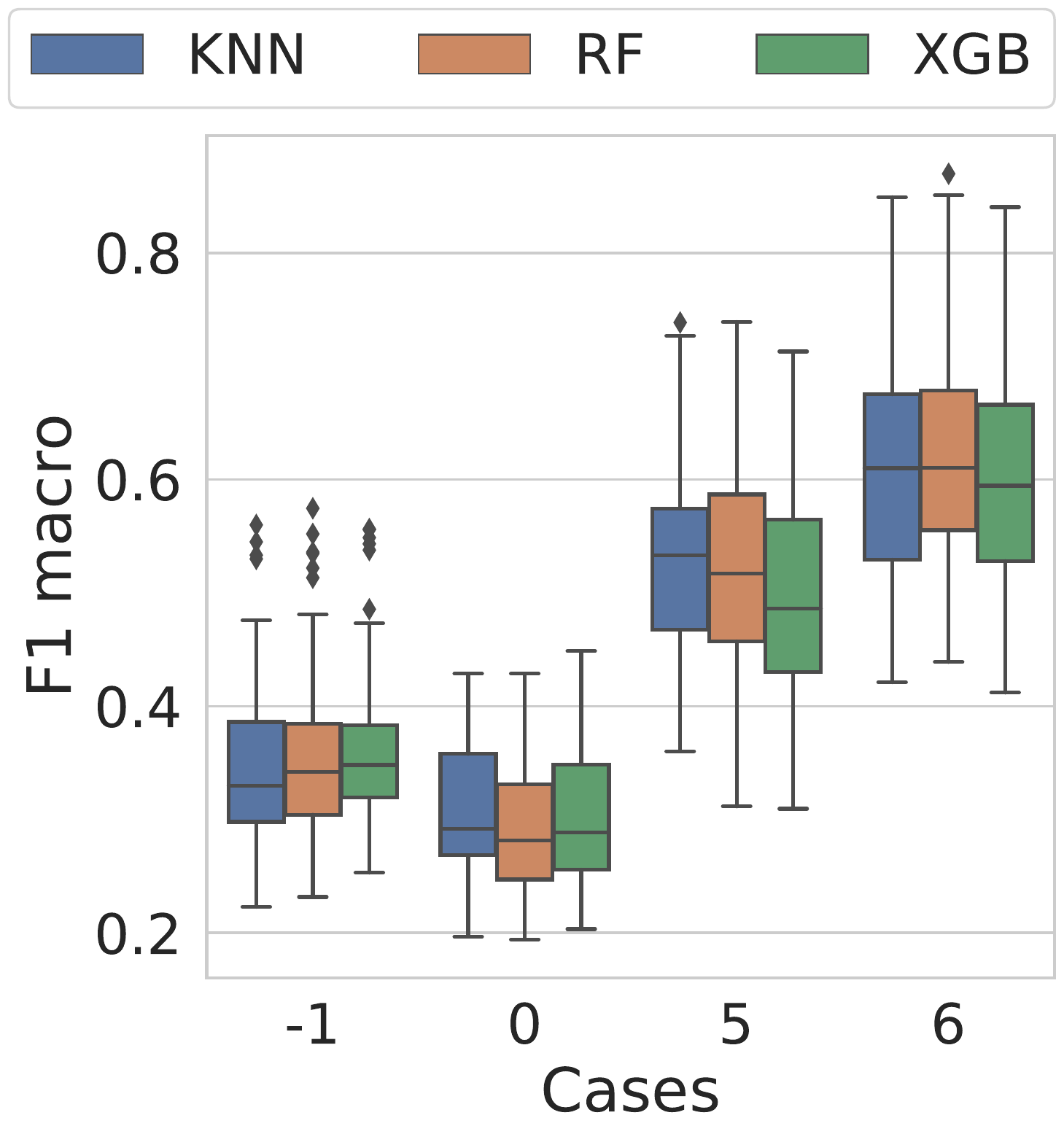}%
\hfill
\includegraphics[width=0.24\textwidth]{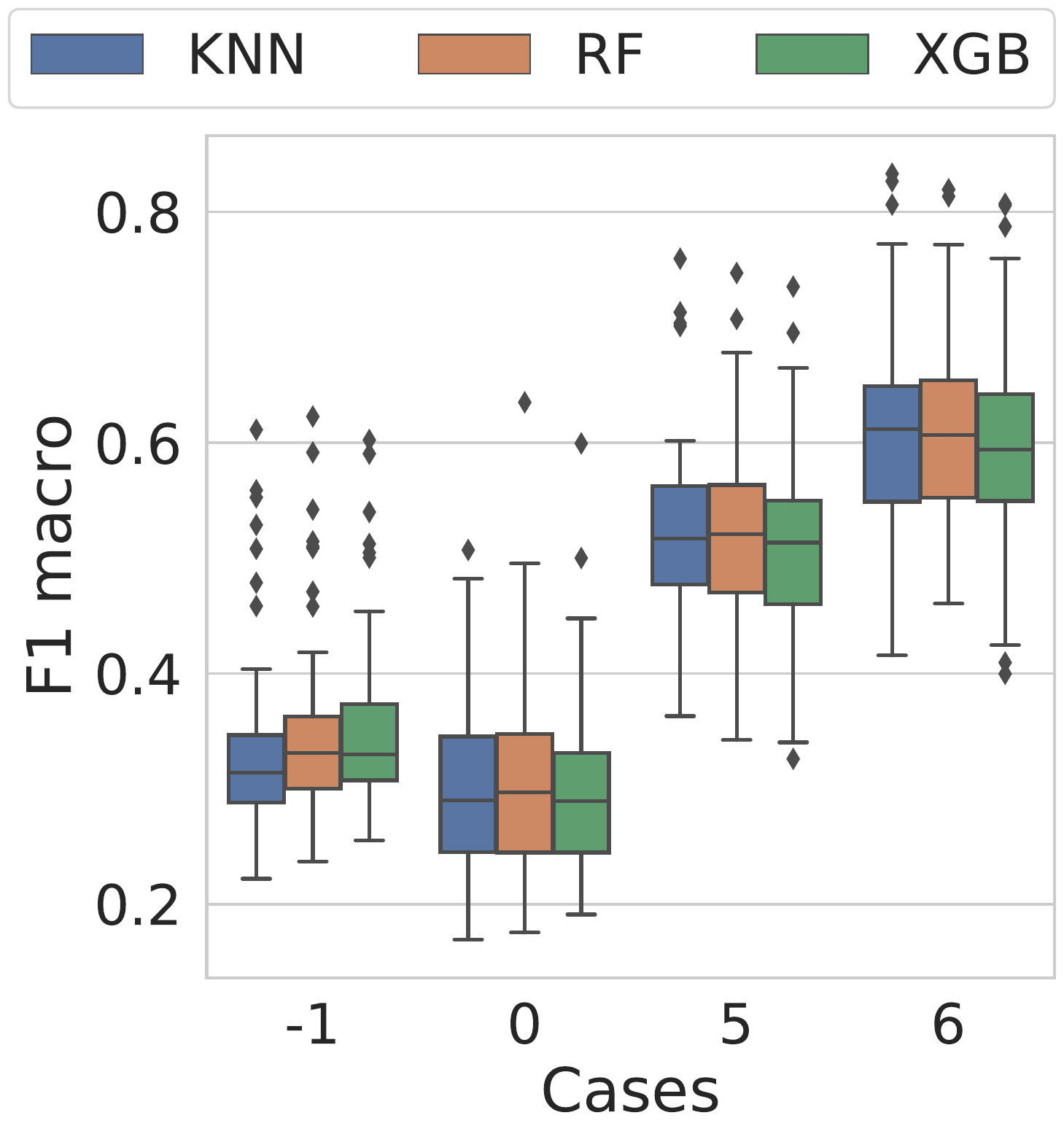}%
\hfill
\includegraphics[width=0.24\textwidth]{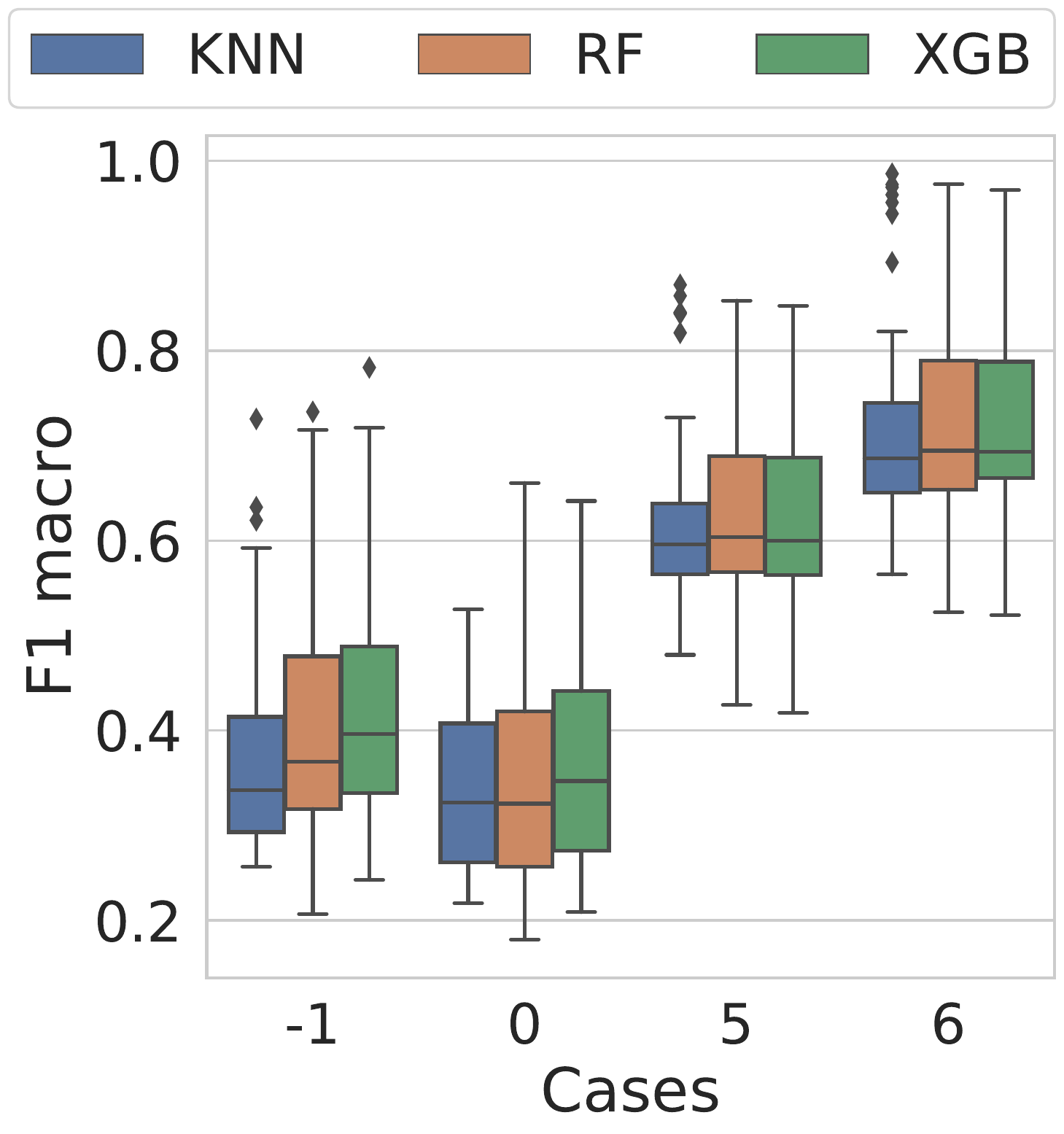}%
\caption{Personalized PLR model performance in investigated cases. From left to right, figures represent results on \textbf{EDA, ECG, EMG,} and \textbf{EDA + ECG + EMG} modalities, respectively. Each point in each box-whisker plot indicates F1-macro computed on each participant. }
\label{fig:person}
\vspace{-5mm}
\end{figure*}

\begin{figure*}
\centering
\includegraphics[width=0.24\textwidth]{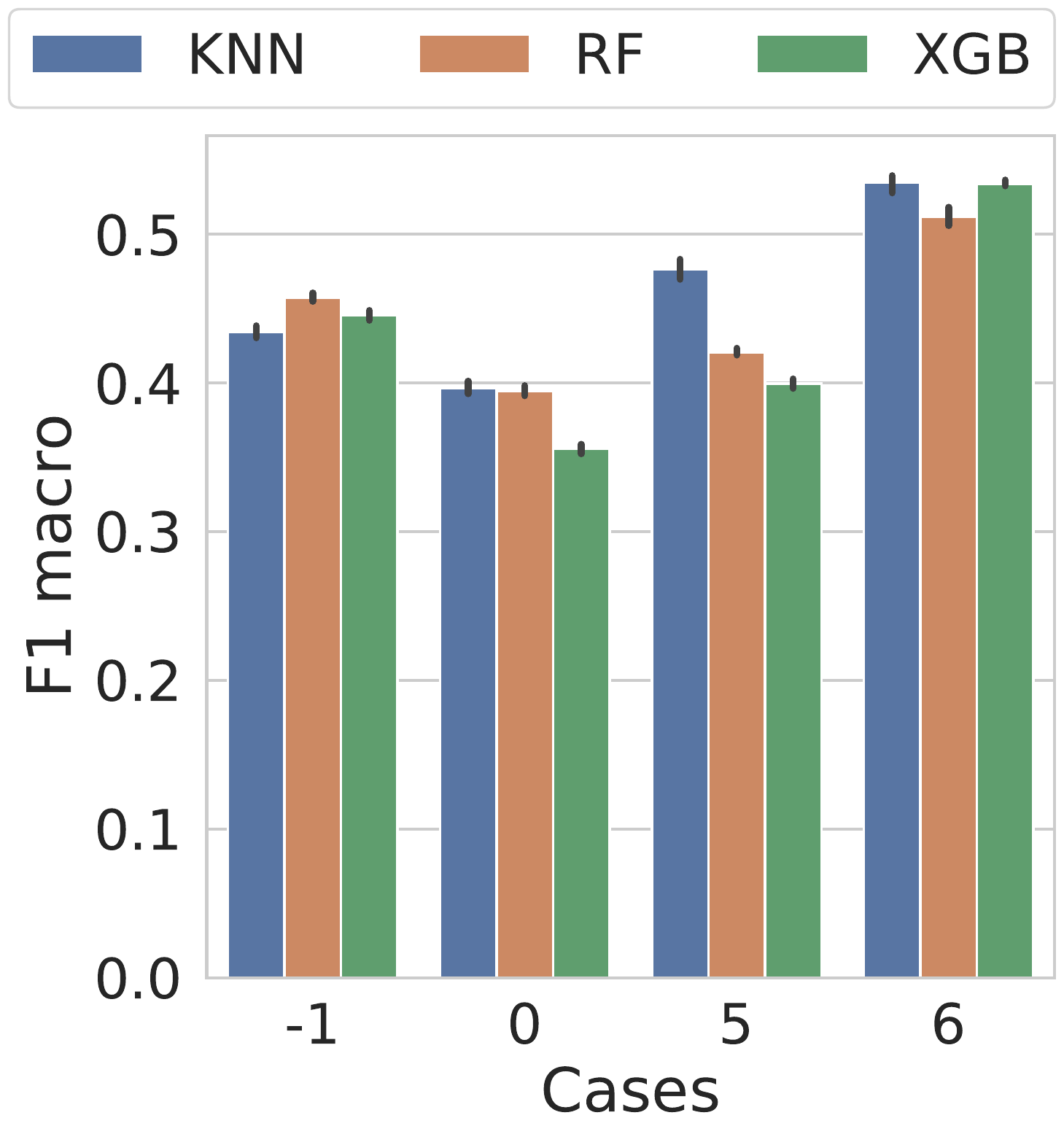}%
\hfill
\includegraphics[width=0.24\textwidth]{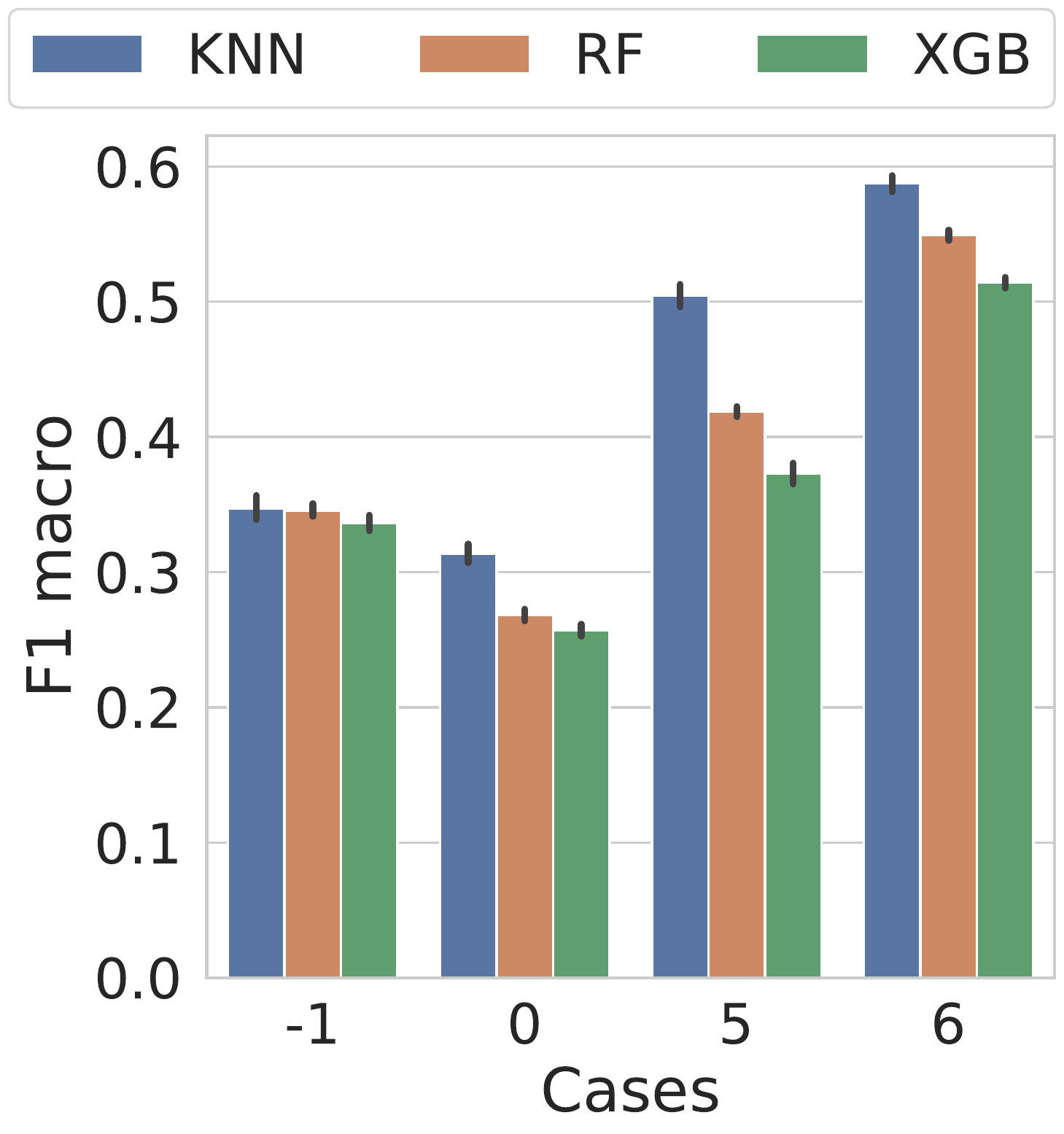}%
\hfill
\includegraphics[width=0.24\textwidth]{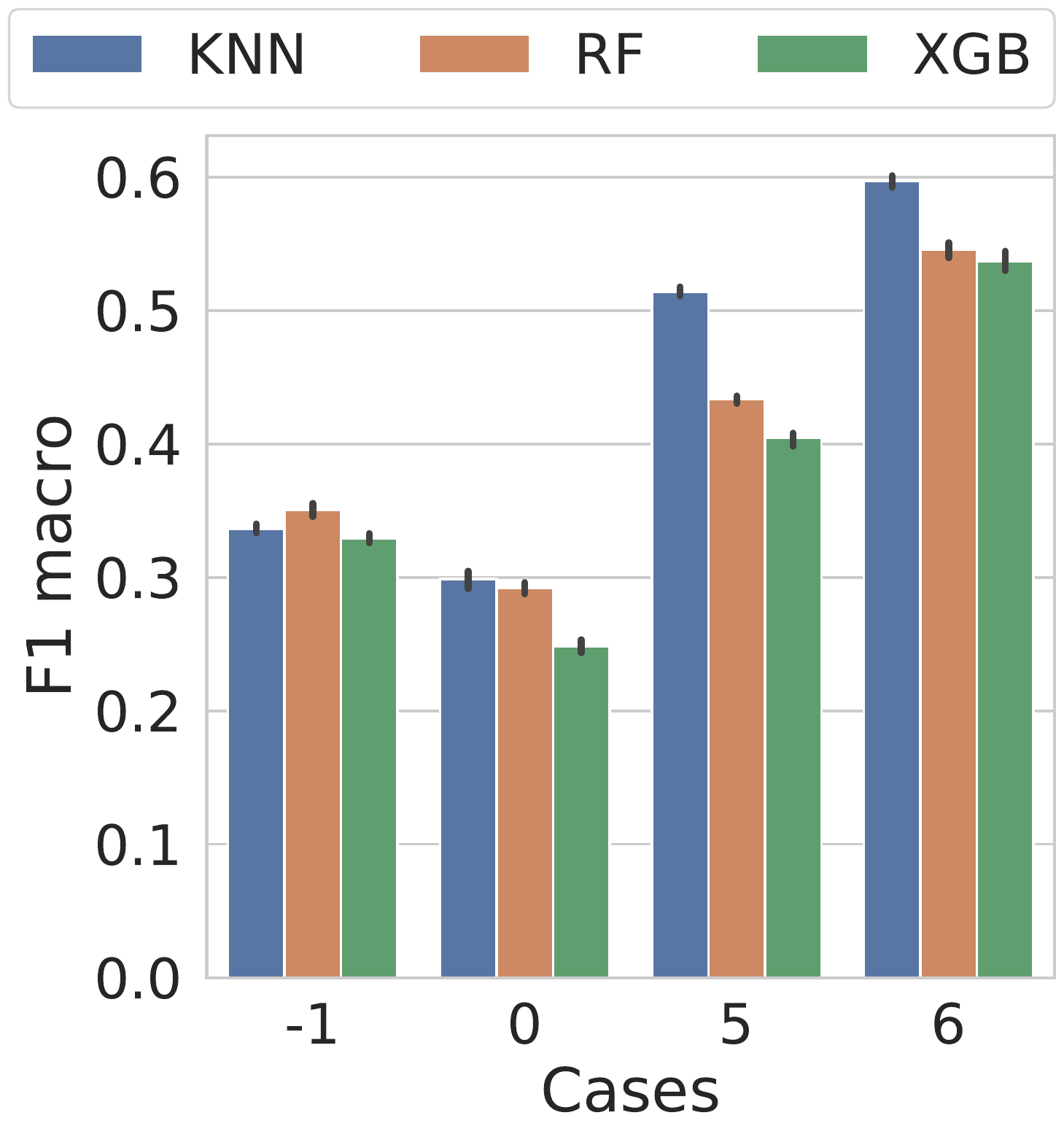}%
\hfill
\includegraphics[width=0.24\textwidth]{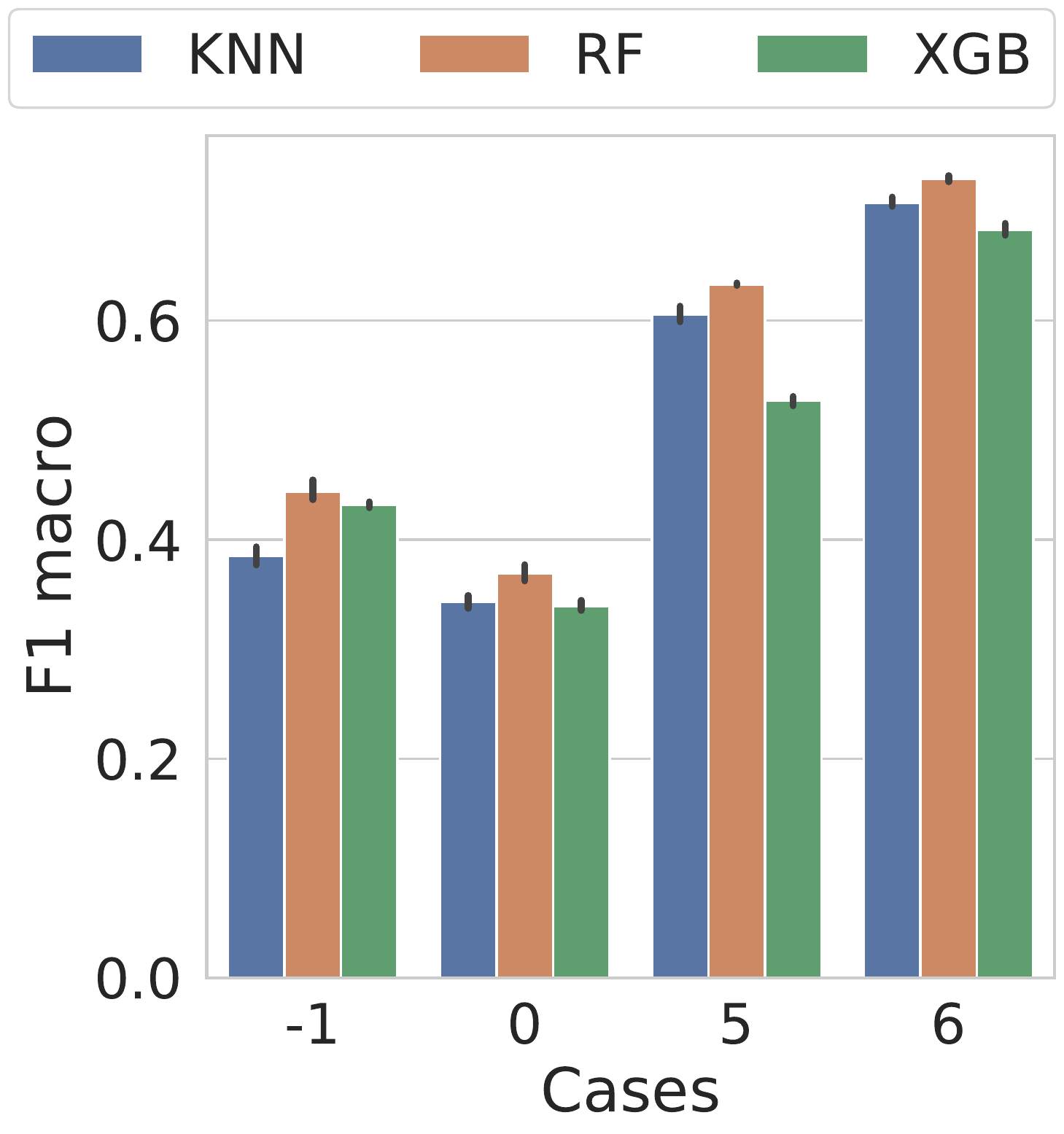}%
\caption{PLR model performance on \textbf{known participants} (we assumed participants are likely to known) in investigated cases. From left to right, figures represent results on \textbf{demographic information with EDA, ECG, EMG,} and \textbf{EDA + ECG + EMG} modalities, respectively. Demographic information was incorporated in feature vector as context. }
\label{fig:genRandWDemog}
\vspace{-5mm}
\end{figure*}

\begin{figure*}
\centering
\includegraphics[width=0.24\textwidth]{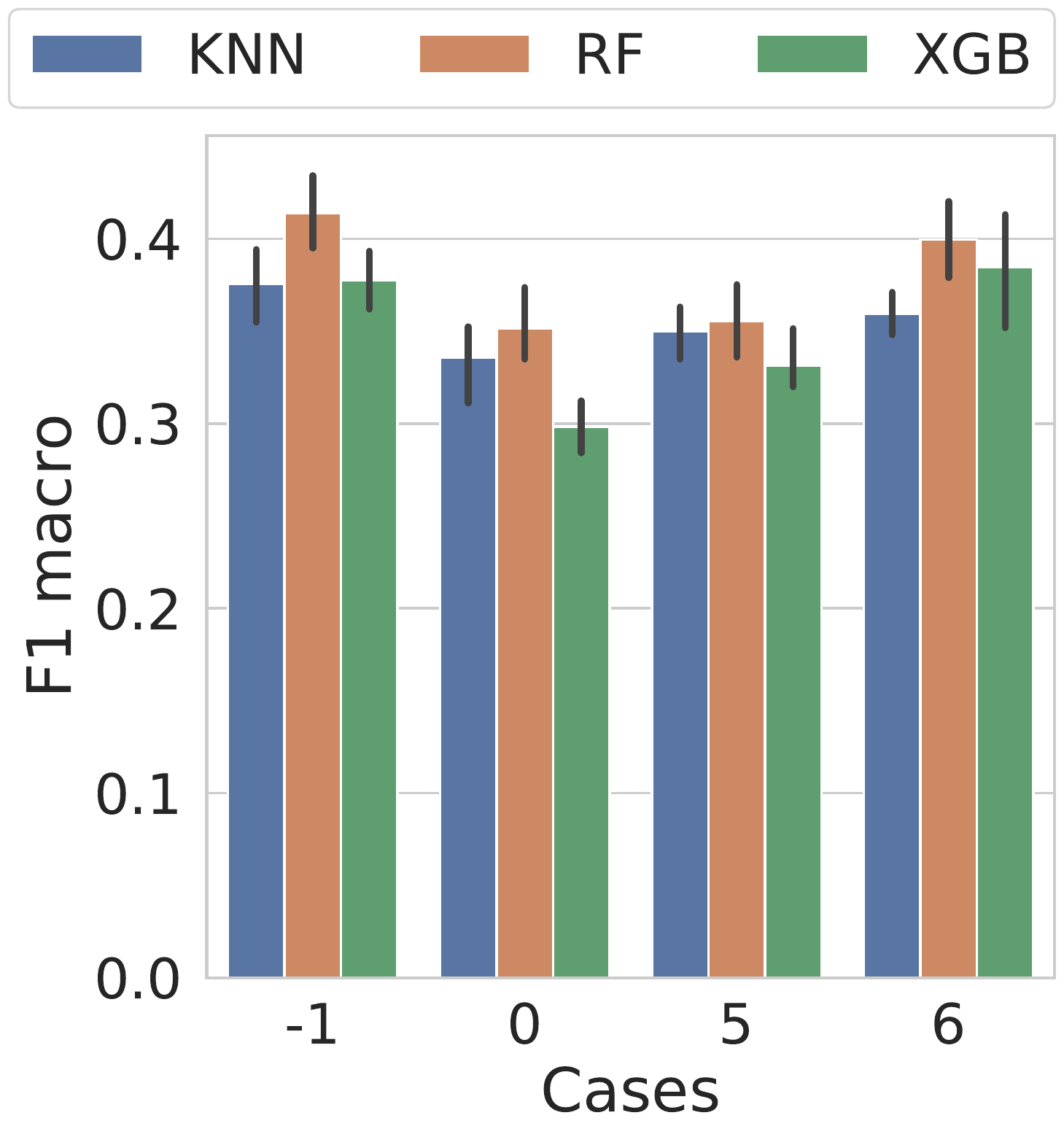}%
\hfill
\includegraphics[width=0.24\textwidth]{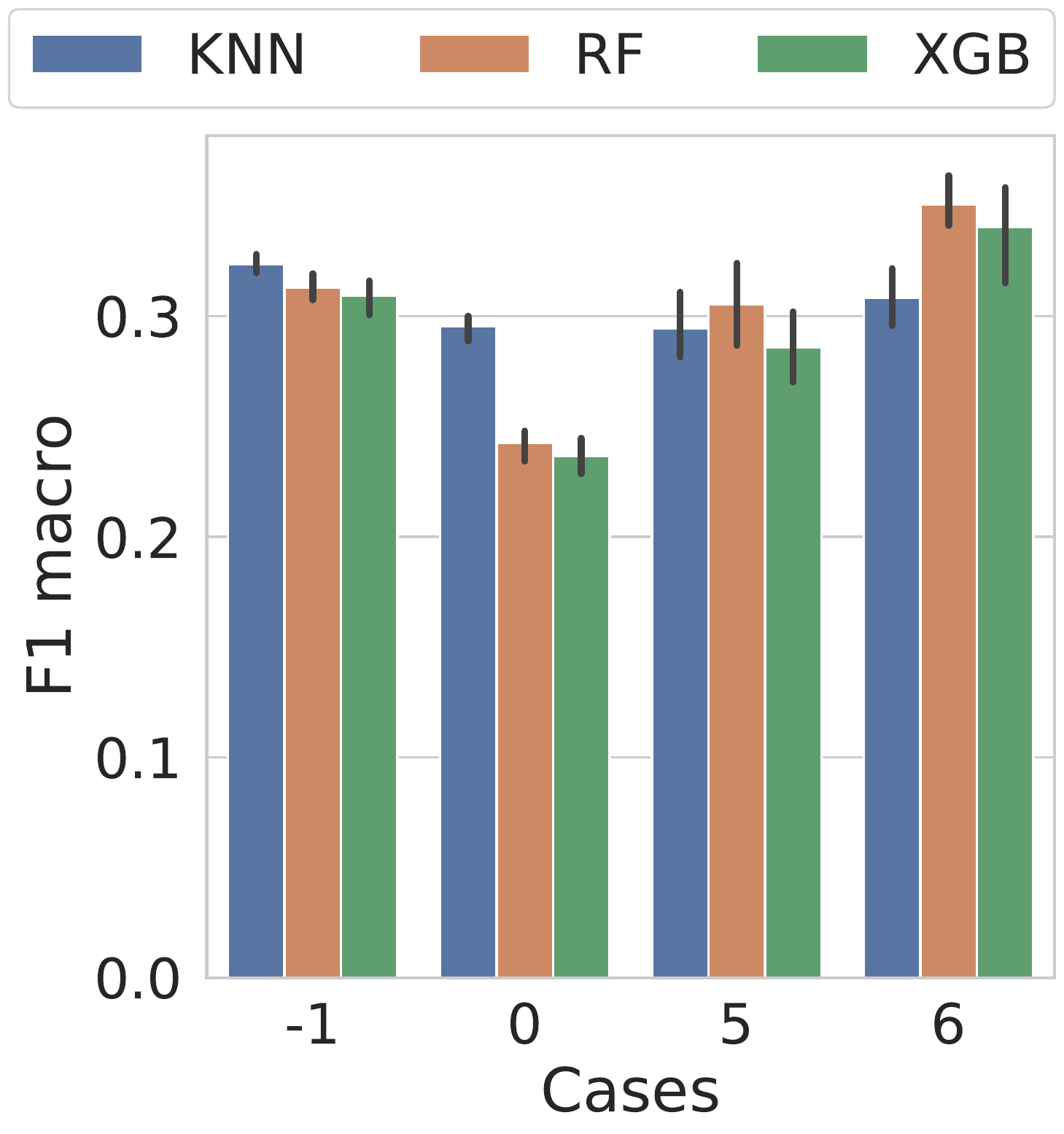}%
\hfill
\includegraphics[width=0.24\textwidth]{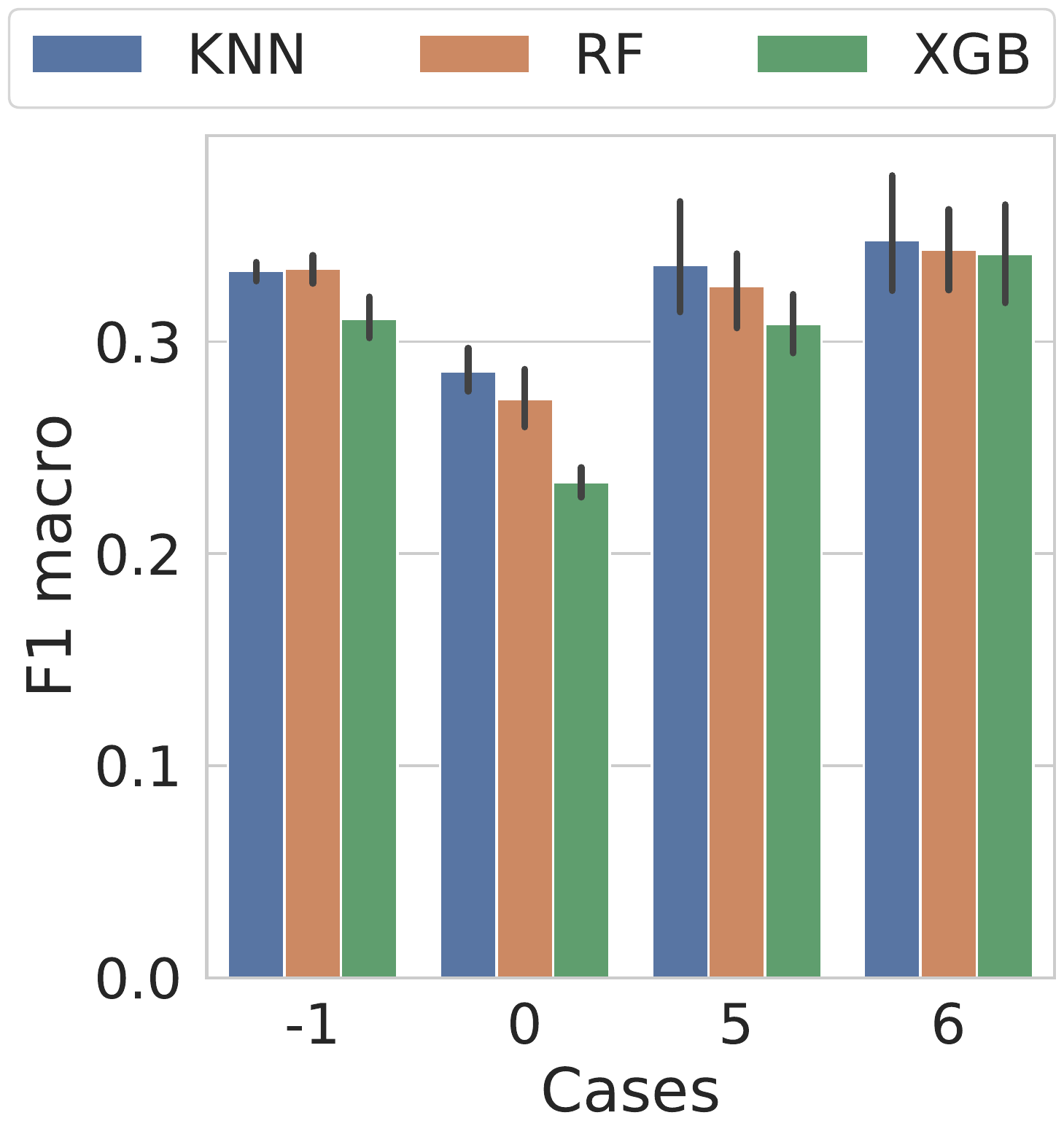}%
\hfill
\includegraphics[width=0.24\textwidth]{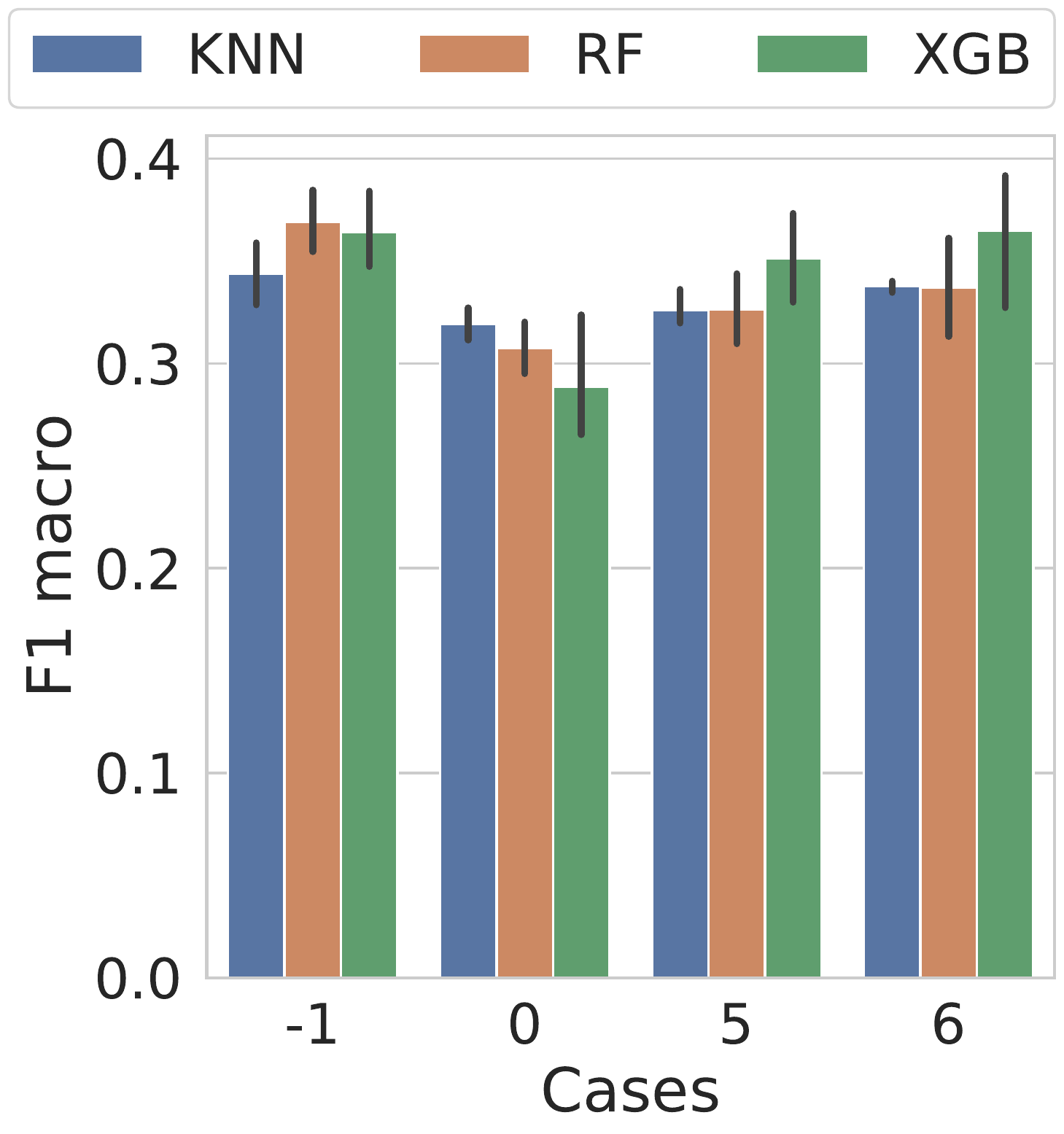}%
\caption{PLR model performance on \textbf{unknown participants} in investigated cases. From left to right, figures represent results on \textbf{demographic information with EDA, ECG, EMG,} and \textbf{EDA + ECG + EMG} modalities, respectively. Demographic information was incorporated in feature vector as context.
}
  \label{fig:genSubOutWDemog}
\vspace{-5mm}
\end{figure*}

\section*{Section 2: Further bioVid Pain-Affect Dataset Details}
\label{sec:dataset_details}

\textbf{Stimulus}. To elicit spontaneous pain, in the bioVid pain dataset, self-calibrated heat was applied as the stimulus. There are five pain levels in the pain dataset including baseline (BL) - pain level (\textbf{PL}) = $0$, pain threshold - $PL = 1$, pain tolerance level - $PL = 4$; after collecting data from the participants, two intermediate pain levels were selected: $PL = 2$ and $PL = 3$. In the bioVid emotion dataset, video clips from movies were used to elicit spontaneous discrete emotions including amusement (Am), anger (An), disgust (Di), fear (F), and sadness (S). 

\textbf{Modalities}. The bioVid emotion dataset contains three biopotential signals: EDA, ECG, EMG of the trapezius muscle, and videos of participants' frontal face. In addition to the above-mentioned modalities, bioVid pain dataset contains EMG signal collected from corrugator and zygomaticus muscles. In the merged bioVid pain-affect dataset, we only selected the common biopotential signals (e.g., EDA, ECG, EMG from trapezius muscle) and videos. In this work, we focus only on the biopotential signals as studies in healthcare \cite{khezri2015reliable,cowen2015assessing} indicated that biopotential signals are major objective indicators of pain and other affect.

\textbf{Sample length}. The bioVid pain dataset contains $5.5$ seconds long subsamples (from the raw data) along with raw data, while the bioVid emotion dataset contains raw data (range = $[32, 245]$ seconds long) only. In this work, we extracted $5.5$ seconds long subsamples from the raw emotion data following the pain dataset to have the same length samples for pain, baseline, and affect in our merged dataset. 

\textbf{Discrete categories}. As pointed out by Werner et al. \cite{werner2017analysis}, the visual responses of PL 1 and PL 2 in bioVid pain dataset were difficult to distinguish due to the similarity in terms of responses and some participants not showing any affective responses. Further, PL 3 and PL 4 had similar responses to a certain extent. Based on those findings, in this new merged dataset, we merged PL 1 and PL 2 to one level named low-level pain (LLP), and PL 3 and PL 4 to another pain level named high-level pain (HLP). Since our primary goal is to investigate the influence of affect in PLR models we merged all discrete affects (e.g., Am, An, Di) to one specific affect class named Affect (A). Hence, in the new bioVid pain-affect dataset, we have four discrete categories: BL, LLP, HLP, and A. 

\section*{Section 4: Further Experiments}

\textbf{Evaluation setup.} Since the major goal of this study is to investigate the limitation of previous PLR model, and evaluate the effectiveness of proposed approach, we performed two additional types of model validations: i) \textbf{evaluation on unknown/new participants}: we used leave-n-participant-out validation, where $n=15$ in our experiments to evaluate all studied cases on participants who are unknown during training;
ii) \textbf{person-specific evaluation}: in this evaluation, we trained and tested on each participant under the assumption that participants are different in terms of eliciting affective behavior, hence, for a given participant's data, we split the data into $70\%$ training and $30\%$ test data, and we ran the experiments $5$ times with different seeds. We used F1 macro (macro-average F1 measure) to report the model performance given its robustness towards balanced and unbalanced dataset.

\textbf{Generalization to unknown, known participants, and personalization.} The following experiments were performed to evaluate whether we could build generalizable PLR model that can generalize to everyone as people are likely to be different in terms of eliciting affects. Comparing Figures \ref{fig:genRand} to \ref{fig:genSubOut}, and Figures \ref{fig:genRandWDemog} to \ref{fig:genSubOutWDemog}, we can see that when we assumed participants are known we are seeing improvement in performance, this could be due to the subjective variability. In Figure \ref{fig:lSpace}, we can see that it is easier to distinguish between B, LLP, HLP, and A in person-specific case, however, when sampled from multiple people (demography-specific, general), it is harder to distinguish. 

An interesting pattern that can be observed is that as we are increasing the number of subjects in samples, it is getting more and more difficult to distinguish between affective states, which in turn likely to make generalization to unknown participants harder. We also looked into demographic information as context to evaluate the influence of gender, age, and demographic group \footnote{Demographic group is constructed via grouping participants based on their age, and gender. Female aged in between $[20, 30)$, $[30, 50)$, and $[50, 65)$ are grouped and labeled as $F_1, F_2, F_3$, respectively. Male in similar age categories are grouped and labeled as $M_4, M_5, M_6$}. The results suggest that demographic information showed some improvement in PL recognition performance in known participants setting as presented in Figure \ref{fig:genRandWDemog}. 

Figure \ref{fig:person} shows that the PLR model gave encouraging results for some participants, however, for others even though model was personalized (trained and tested on each participant), it is challenging to distinguish between multiple states. There could be multiple explanations for this. First, it is possible those subjects did not feel affect or pain during experiments. Secondly, they may have been going through some other affective states that were not captured in the data. Lastly, the physiological data was not different across the affective states for those participants (i.e. pain and affect looked similar). Considering this, we need to further look into the data to make any concluding remarks. Finally, in all evaluations (unimodal and multimodal features with/without incorporation of context), we observed improvement when we accounted for affect in PLR models compared to cases when affects were not accounted for. 

Based on our experiments, we can opine that the effective and reliable way to solve the problem is to incorporate affect in PL recognition approaches. As we can see in Figure \ref{fig:lSpace}, pain level recognition when accounting for affect is a challenging task due to the similarity of data patterns of pain and non-pain affect, as well as the variability of data based on people differences \cite{hu2016painful}. The proposed formulation still performed comparatively better due to its ability to adapt to affect dataset. Hence, even though incorporation of affect could make the modeling challenging, it would make the modeling sound, reliable, and realistic.

\end{document}